\documentclass[final]{egpubl}
\makeatletter 
\providecommand{\p@copyrightTextShort}{} 
\providecommand{\p@copyrightTextShortEven}{} 
\providecommand{\p@copyrightTextTitPag}{} 
\makeatother 

\ConferencePaper
\usepackage[T1]{fontenc}
\usepackage{dfadobe}  

\usepackage{cite}

\usepackage[pdftex]{graphicx} \pdfcompresslevel=9

\usepackage{lineno}

\usepackage{xcolor}
\usepackage{subcaption}
\usepackage{dsfont}
\usepackage{booktabs}
\usepackage{amsmath}
\usepackage{amssymb}
\usepackage{pifont}
\usepackage{caption}
\usepackage{hyperref}
\usepackage[utf8]{inputenc} 

\newcommand{\rev}[1]{{\color{black} #1}}

\usepackage[font=it]{caption}
\title[\rev{VQ-Style:} Disentangling Style and Content in Motion with Residual Quantized Representations]{\rev{VQ-Style:} Disentangling Style and Content in Motion with Residual Quantized Representations}

\author[Zargarbashi et. al.]
{\parbox{\textwidth}{\centering Fatemeh Zargarbashi$^{1,2}$, Dhruv Agrawal$^{1,2}$, Jakob Buhmann$^{2}$, Martin Guay$^{2}$, Stelian Coros$^{1}$, Robert W. Sumner$^{1,2}$} 
\\
\centering $^1$ETH Z\"urich, Switzerland \quad
         $^2$DisneyResearch|Studios, Switzerland
}

\begin{document}

\teaser{ 
 \includegraphics[width=\linewidth]{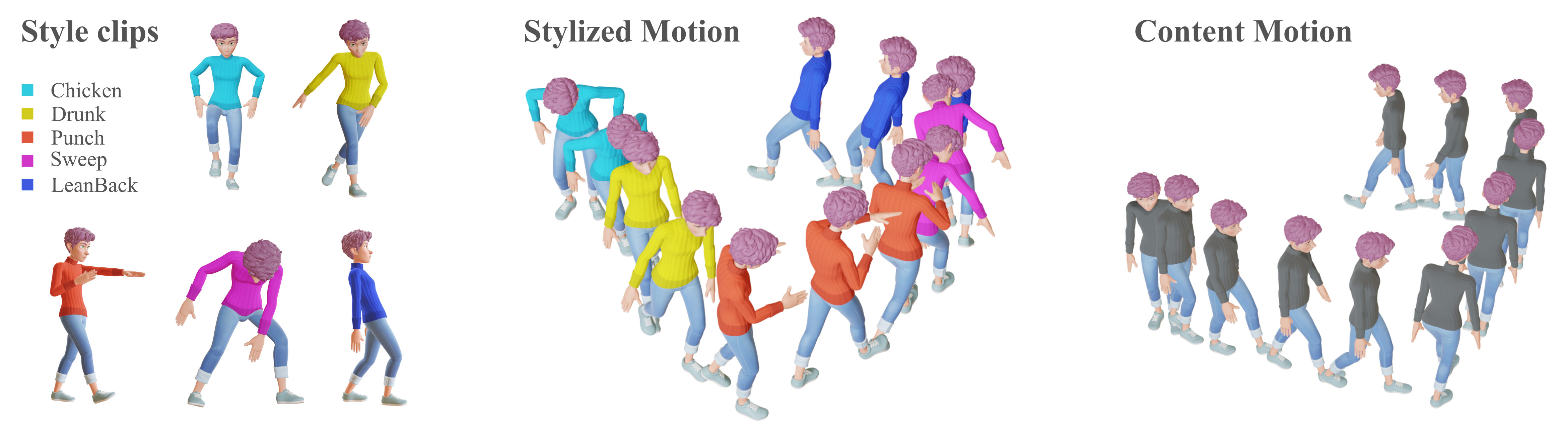}
 \centering
  \caption{On the left, we see different styles, while at the far right is a content motion that is much longer.  In the middle, we transferred the style from different clips over different time windows of the content while seamlessly transitioning between them.}
\label{fig:teaser}
}

\maketitle

\begin{abstract}
Human motion data is inherently rich and complex, containing both semantic content and subtle stylistic features that are challenging to model. 
We propose a novel method for effective disentanglement of the style and content in human motion data to facilitate style transfer. Our approach is guided by the insight that content corresponds to coarse motion attributes while style captures the finer, expressive details. 
To model this hierarchy, we employ Residual Vector Quantized Variational Autoencoders (RVQ-VAEs) to learn a coarse-to-fine representation of motion. 
We further enhance the disentanglement by integrating codebook learning with contrastive learning and a novel information leakage loss to organize the content and the style across different codebooks. 
We harness this disentangled representation using our simple and effective inference-time technique \emph{Quantized Code Swapping}, which enables motion style transfer without requiring any fine-tuning for unseen styles. 
Our framework demonstrates strong versatility across multiple inference applications, including style transfer, style removal, and motion blending.



\end{abstract}  

\newcommand{\FigFramework}{
    \begin{figure*}[tb]
        \centering
        \includegraphics[width=\linewidth]{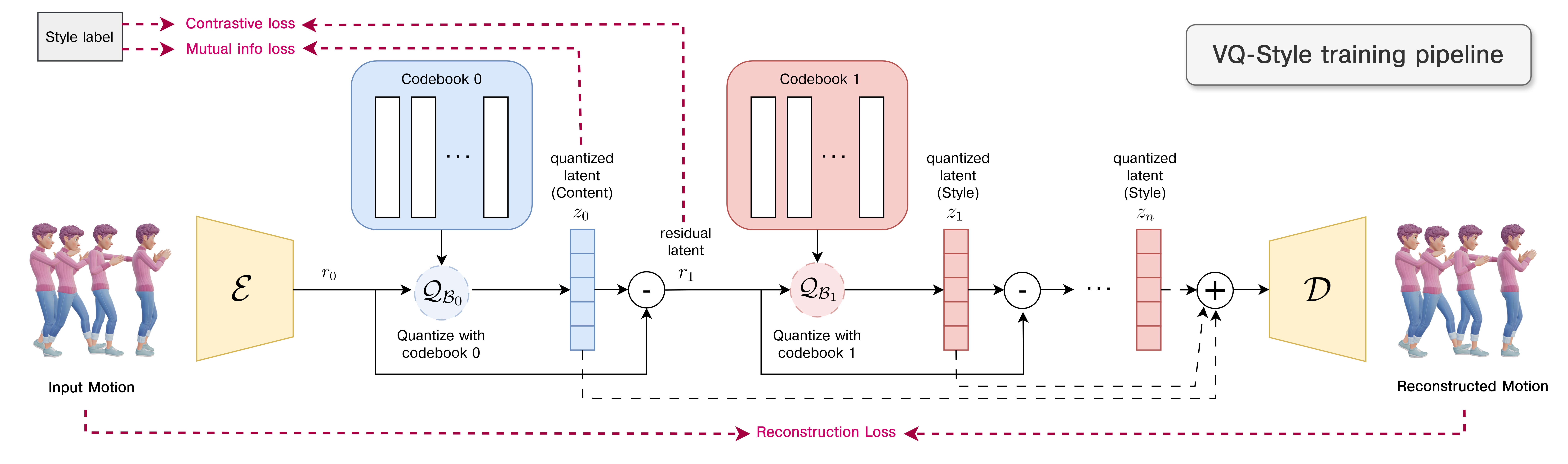}
        \caption{\rev{Our approach encodes motion into a number of codebooks stacked in a residual manner.  Our training strategy enables the content to be represented by the first (blue) codebook, while style is represented by the downstream codebooks (in red).}
        }
        \label{fig:framework}
    \end{figure*}
}

\newcommand{\FigFrameworkStyleTransfer}{
    \begin{figure*}[tb]
        \centering
        \includegraphics[width=0.95\linewidth]{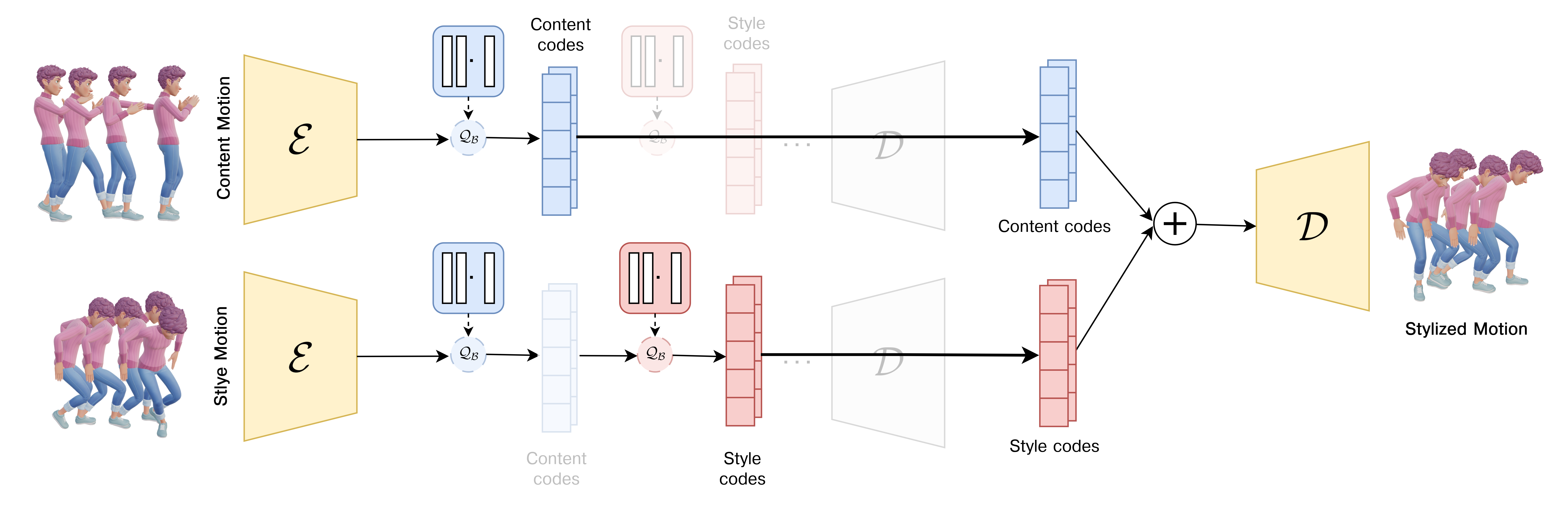}

        \caption{\rev{Quantized Latent Code Swapping. To transfer style from one clip onto another, we encode both clips, and decode a combined embedding constructed by adding the content code (blue) from the content clip and the style code (red) from the second clip.}
        }
        \label{fig:framework_style_transfer}
    \end{figure*}
}

\newcommand{\FigLatentVis}{
    \begin{figure}
        \centering
        \begin{subfigure}{\linewidth}
        \includegraphics[width=\linewidth]{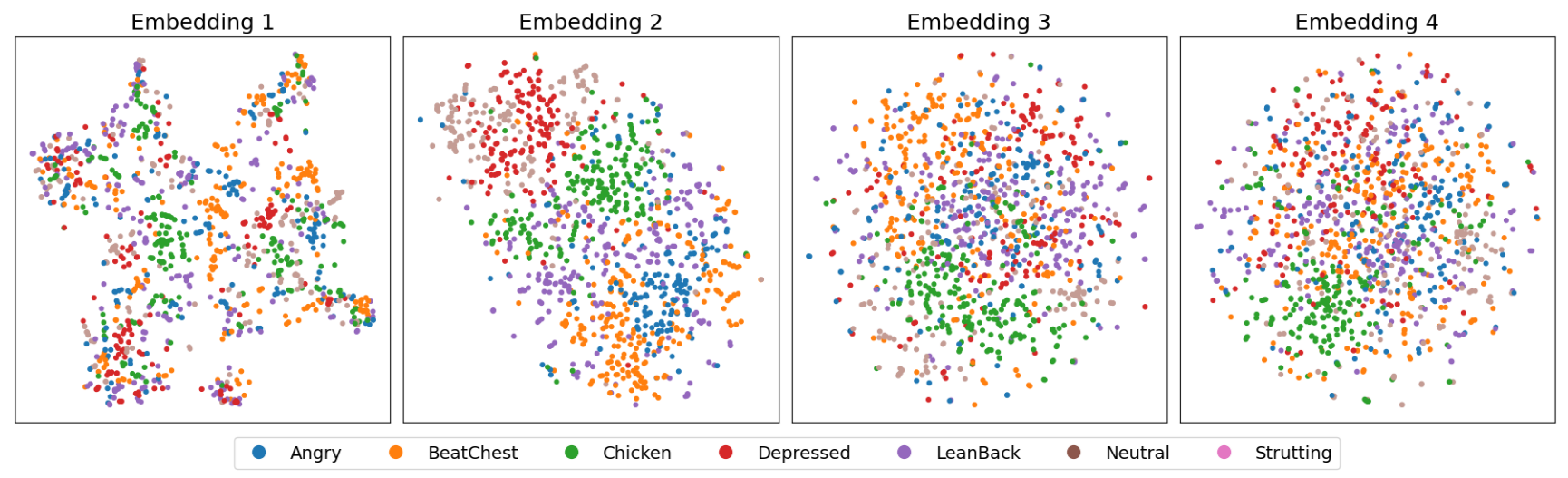}
        \caption{RVQ trained for reconstruction.}
        \label{fig:latent_vis_a}
        \end{subfigure}
        \begin{subfigure}{\linewidth}
            \includegraphics[width=\linewidth]{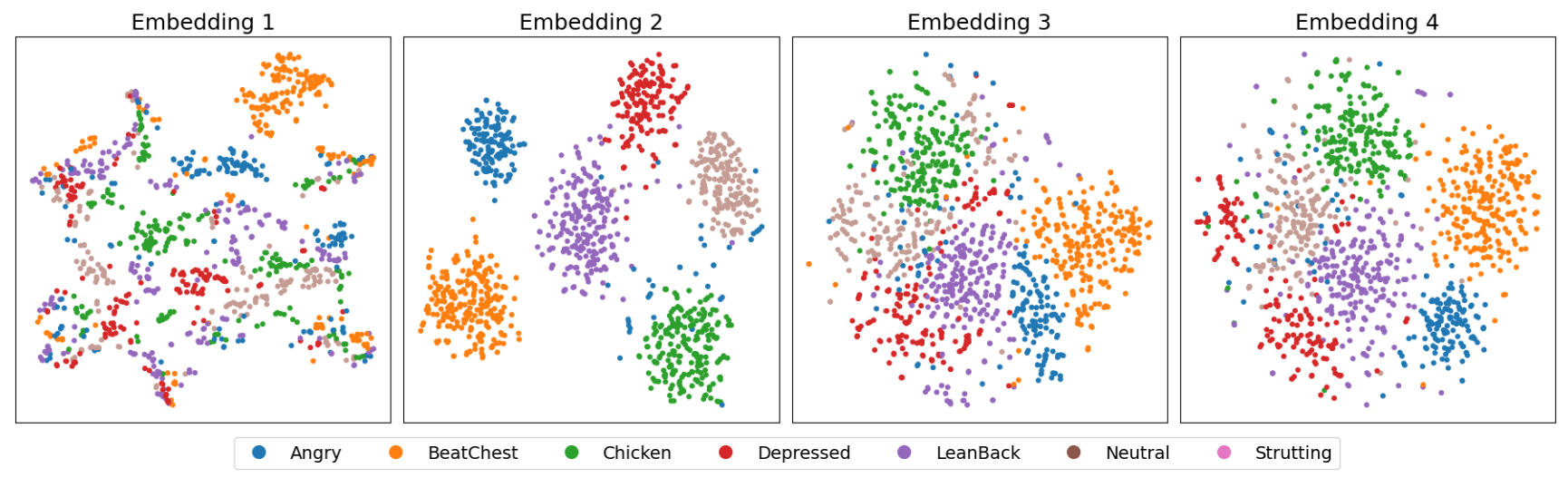}
        \caption{With contrastive learning}
        \label{fig:latent_vis_b}
        \end{subfigure}
        \begin{subfigure}{\linewidth}
            \includegraphics[width=\linewidth]{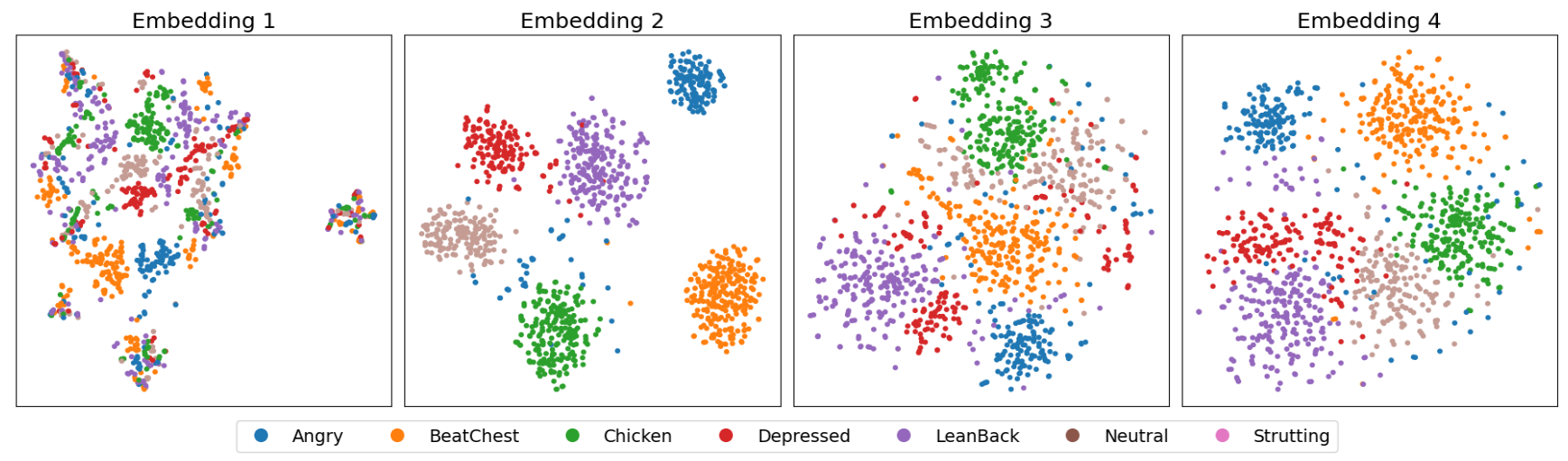}
        \caption{With contrastive learning and mutual info loss.}
        \label{fig:latent_vis_c}
        \end{subfigure}
        \begin{subfigure}{\linewidth}
            \includegraphics[width=\linewidth]{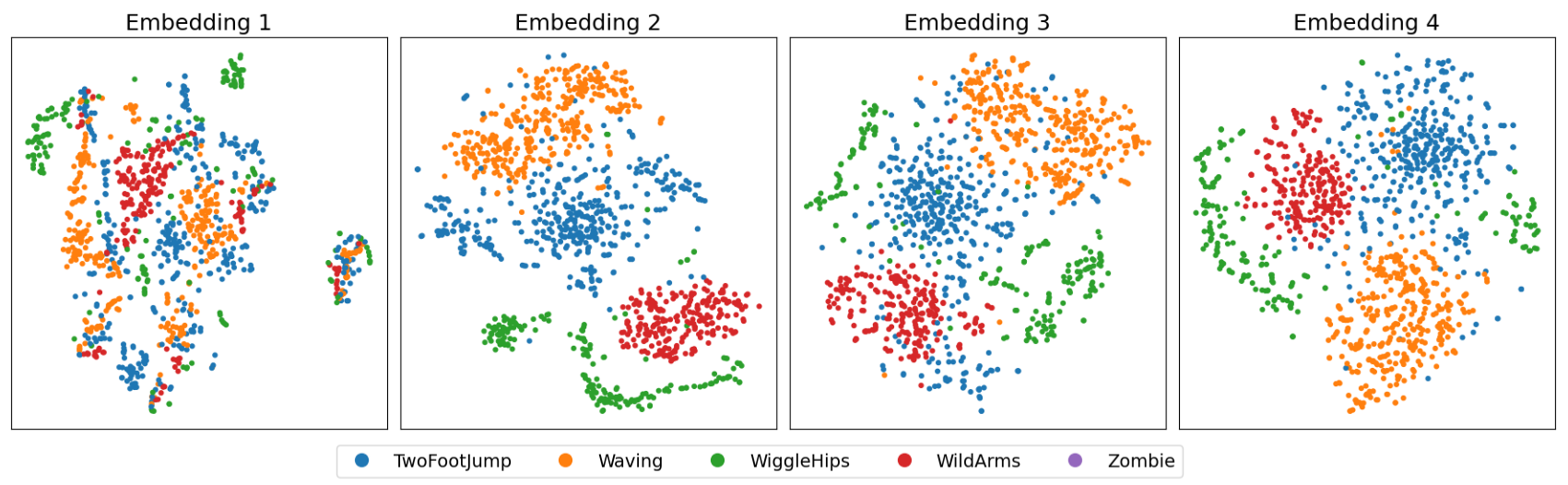}
        \caption{With contrastive learning and mutual info loss (Unseen styles).}
        \label{fig:latent_vis_d}
        \end{subfigure}
        \caption{TSNE plots of residual embeddings illustrating style-content disentanglement in latent space. The colors indicated different style labels.
        \textbf{(a)} Without contrastive learning, styles begin to weakly cluster from the second codebook onward. Embedding 1 remains unclustered, as intended. 
        \textbf{(b)} With contrastive learning, the separation between styles becomes more pronounced. 
        \textbf{(c)} With mutual information loss the disentanglement is further enhanced. 
        \textbf{(d)} The model also disentangles styles never seen during training, demonstrating its generalization ability.
        }
        \label{fig:latent_vis}
    \end{figure}
}

\newcommand{\FigInterpolate}{
    \begin{figure}
        \centering
        \includegraphics[width=\linewidth]{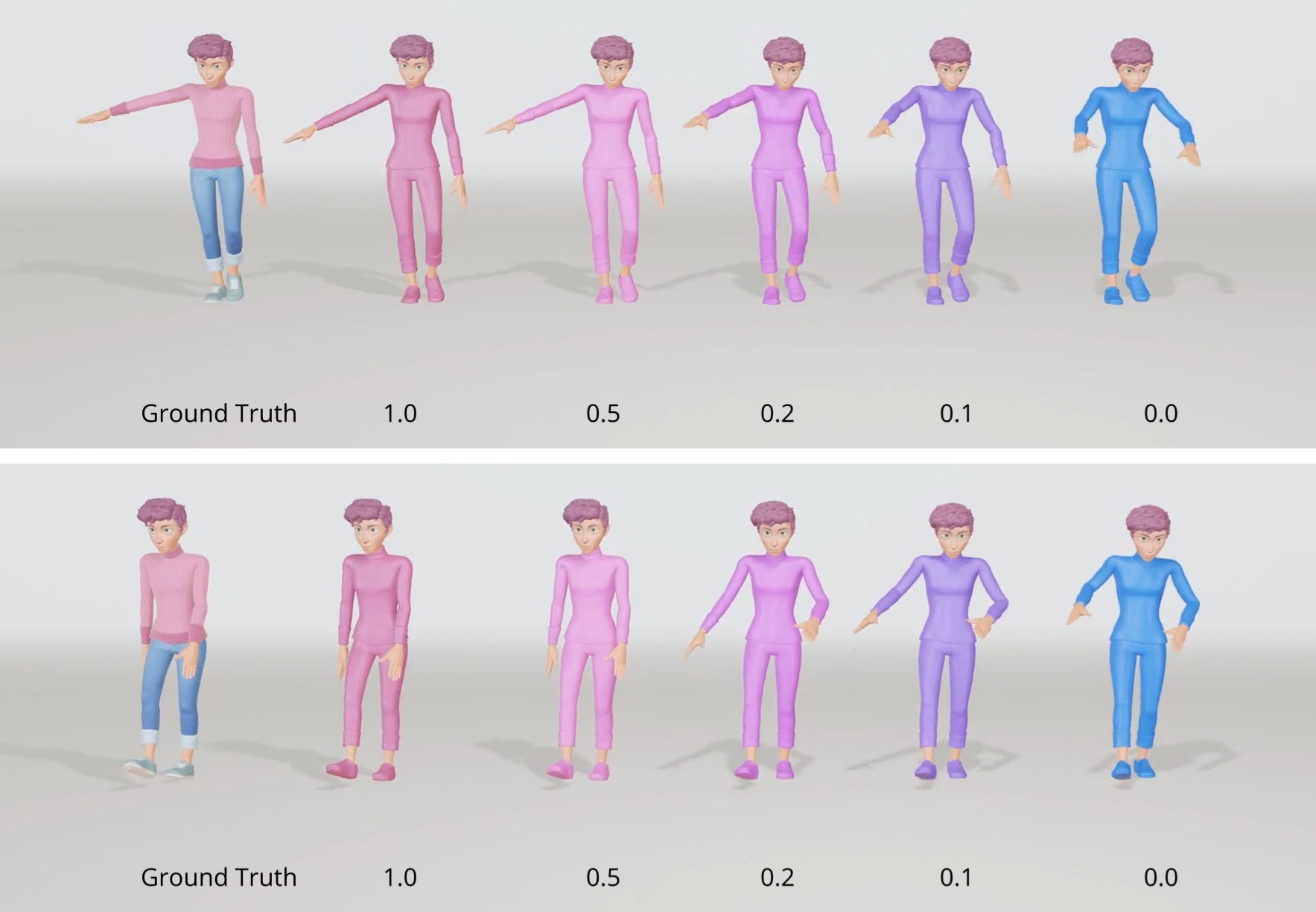}
        \caption{Interpolation between content and stylized motion. Going from left to right, more of the style code is removed. The resulting motion smoothly approaches \emph{Neutral} motion.}
        \label{fig:interpolate}
    \end{figure}
}

\newcommand{\FigContentOnly}{
    \begin{figure*}[tb]
        \centering
        \begin{subfigure}[b]{\textwidth}
            \includegraphics[width=\linewidth]{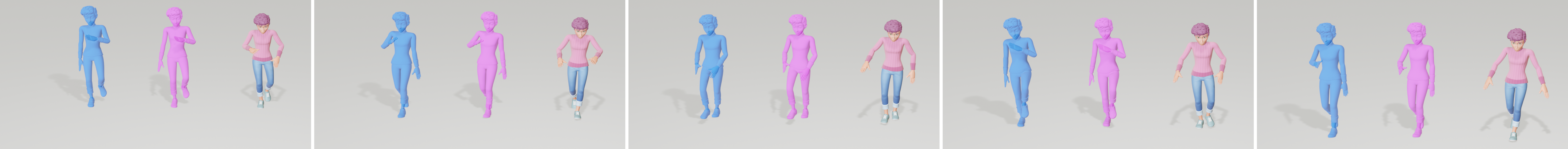}
            \caption{Style: March}
        \end{subfigure}
        \hfill
        \begin{subfigure}[b]{\textwidth}
            \includegraphics[width=\linewidth]{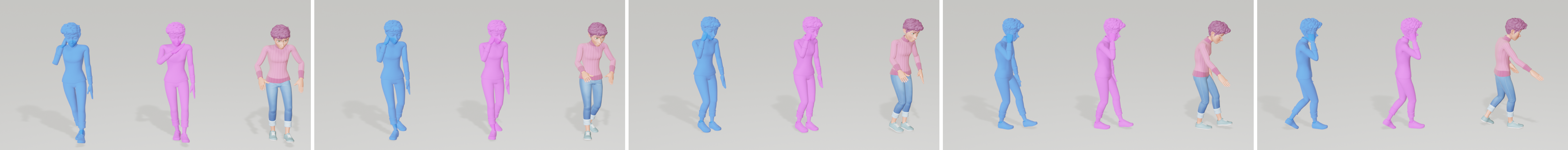}
            \caption{Style: On the phone right}
        \end{subfigure}
        \caption{Content extraction by decoding motion using only the first codebook. \textbf{Left (blue):} ground truth, \textbf{Middle (pink):} full reconstruction, \textbf{Right (texture):} extracted content. Best viewed in color.}
        \label{fig:content_only}
    \end{figure*}
    }

\newcommand{\FigStyleTransferTrain}{\begin{figure}
    \centering
    \begin{subfigure}[b]{\linewidth}
         \centering
         \includegraphics[width=\textwidth]{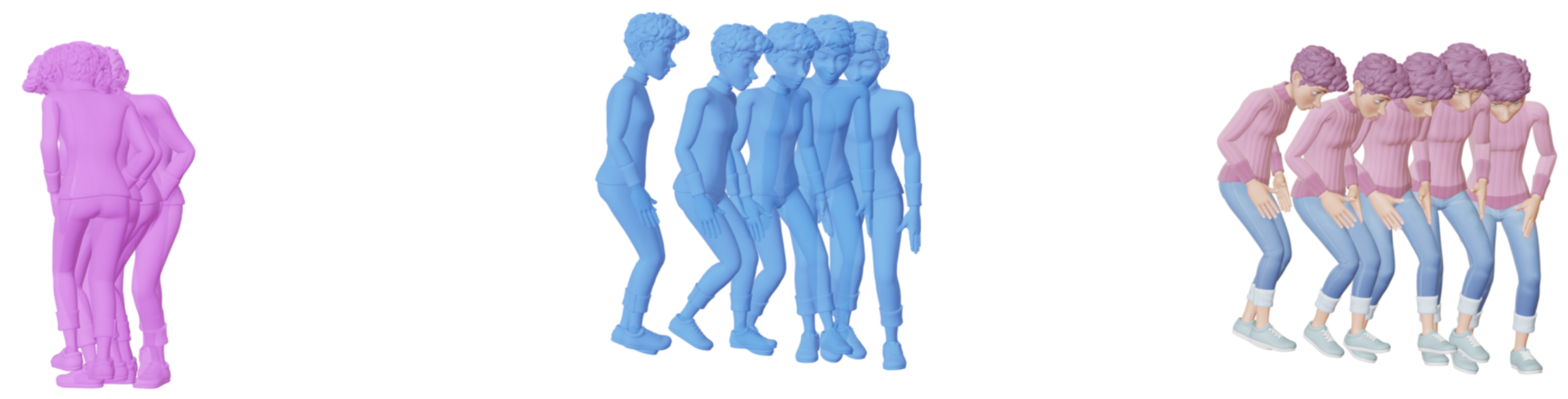}
     \caption{Style: \emph{Old}}
     \end{subfigure}
     \hfill
     \vspace{0.05cm}
     \begin{subfigure}[b]{\linewidth}
     \centering
     \includegraphics[width=\textwidth]{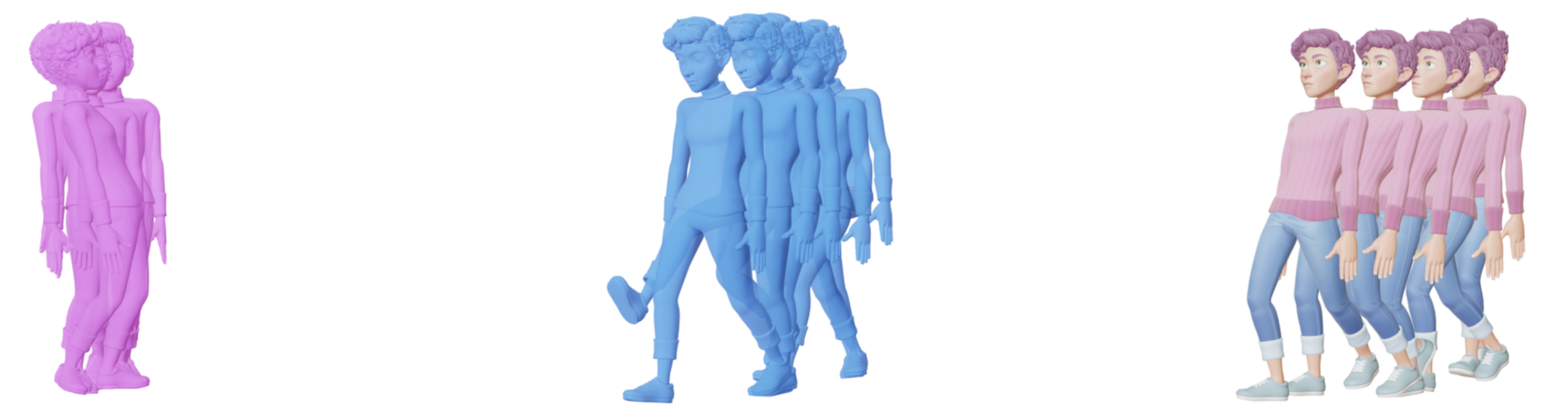}
     \caption{Style: \emph{Leanback}}
    \end{subfigure}
    \caption{Style transfer results on \emph{100STYLES}. \textbf{Left:} Style clip, \textbf{Middle:} Content clip, \textbf{Right:} Style transfer result.}
    \label{fig:style-transfer-train}
\end{figure}
}

\newcommand{\FigStyleTransferTest}{
    \begin{figure}
        \centering
        \begin{subfigure}[b]{\linewidth}
                 \centering
        \includegraphics[width=\textwidth]{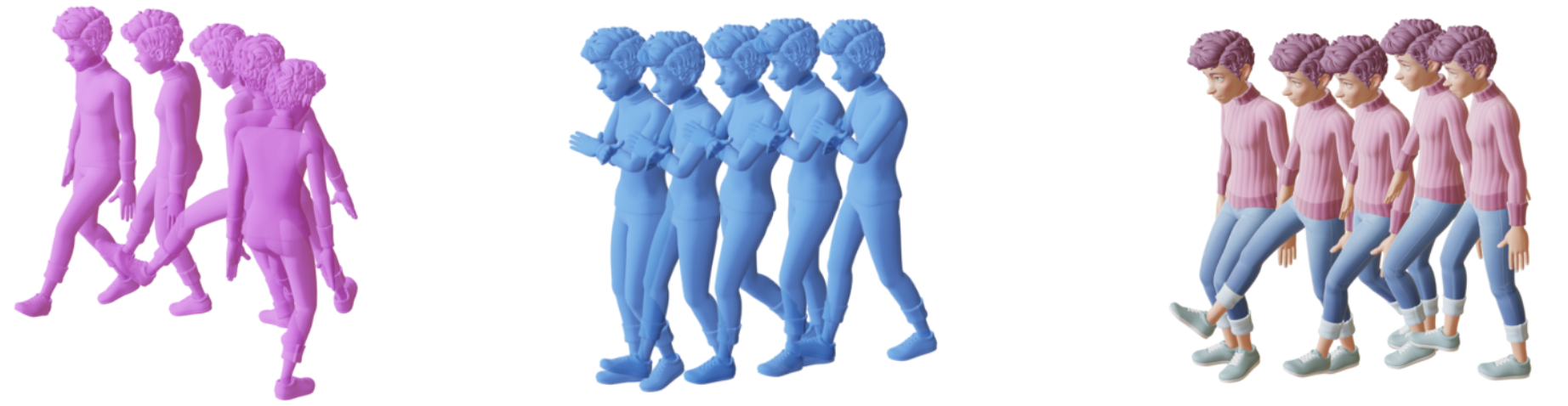}
        \caption{Style: \emph{WildLegs}}
        \end{subfigure}
        \hfill
        \vspace{0.05cm}
        \begin{subfigure}[b]{\linewidth}
        \includegraphics[width=\textwidth]{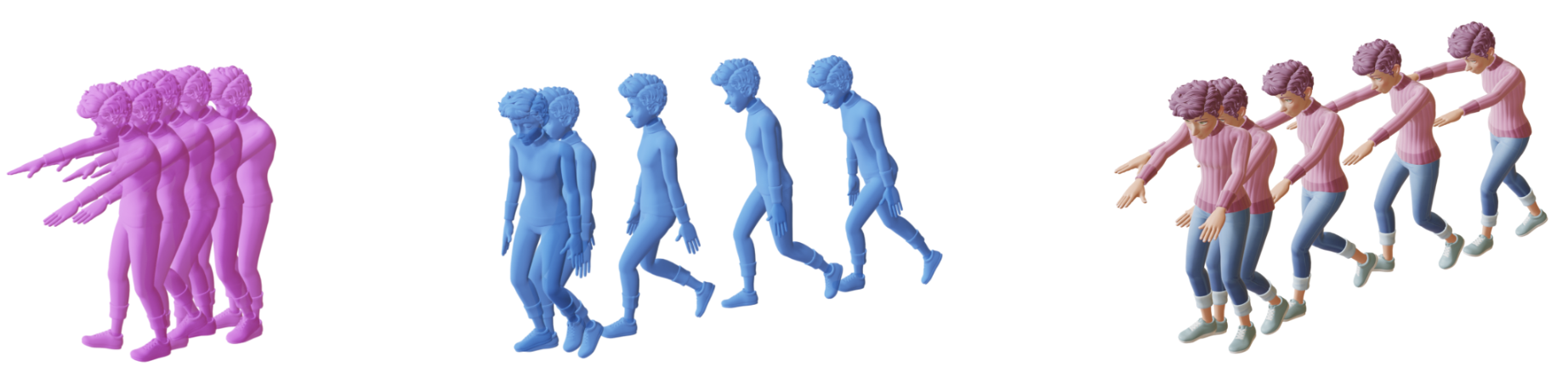}
        \caption{Style: \emph{Zombie}}      
        \end{subfigure}
        \caption{Style transfer for \textbf{Unseen} styles. \textbf{Left:} Style clip, \textbf{Middle:} Content clip, \textbf{Right:} Style transfer result.}
        \label{fig:style-transfer-test}
    \end{figure}}

\newcommand{\FigStyleTransferAberman}{
    \begin{figure*}
        \begin{subfigure}{0.33\linewidth}
            \includegraphics[width=\linewidth, trim = 0 0.5cm 0 0.3cm]{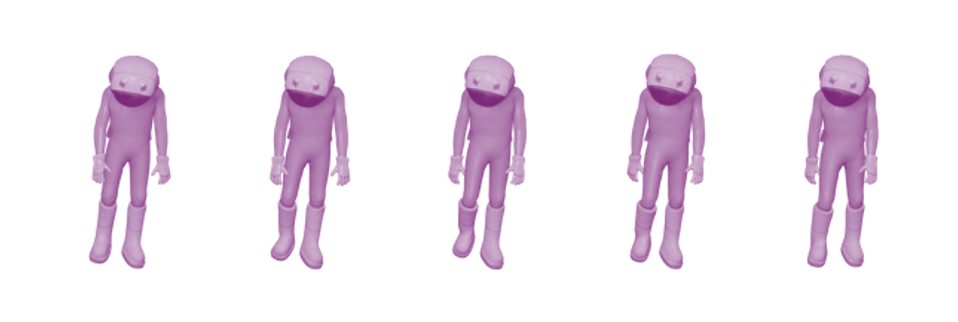}
            \caption*{}
        \end{subfigure}%
        \begin{subfigure}{0.33\linewidth}
            \includegraphics[width=\linewidth, trim = 0 0.5cm 0 0.3cm]{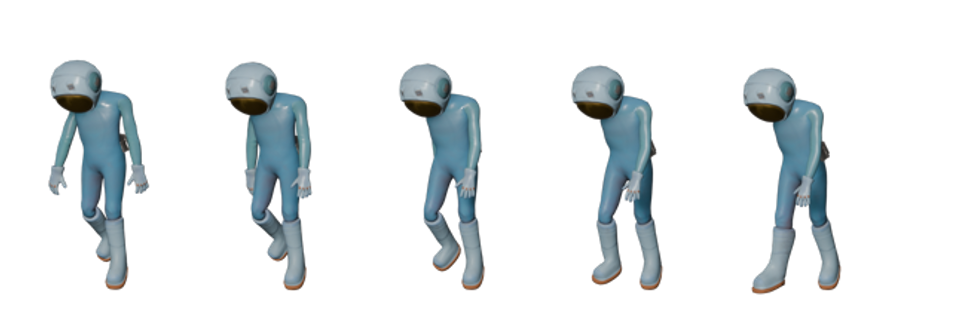}
            \caption{Depressed}
        \end{subfigure}%
        \begin{subfigure}{0.33\linewidth}
        \includegraphics[width=\linewidth, trim = 0 0.5cm 0 0.3cm]{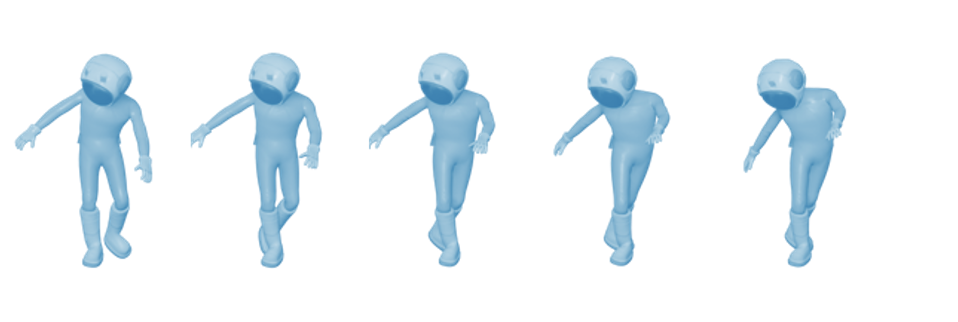}
                    \caption*{}
        \end{subfigure}
        
        \begin{subfigure}{0.33\linewidth}
            \includegraphics[width=\linewidth, trim = 0 0.5cm 0 0.3cm]{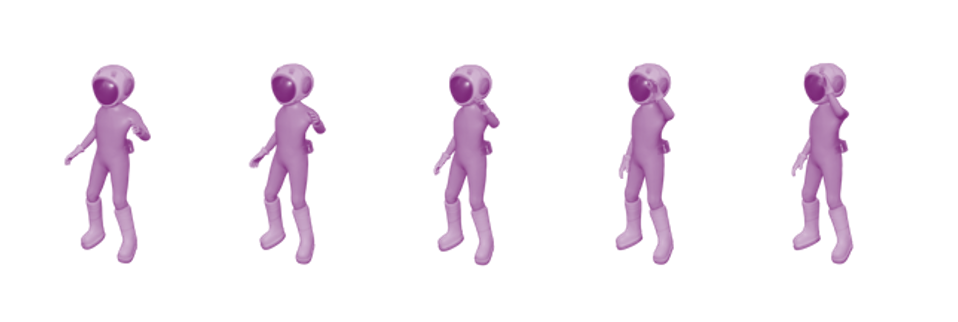}
                    \caption*{}
        \end{subfigure}%
        \begin{subfigure}{0.33\linewidth}
            \includegraphics[width=\linewidth, trim = 0 0.5cm 0 0.3cm]{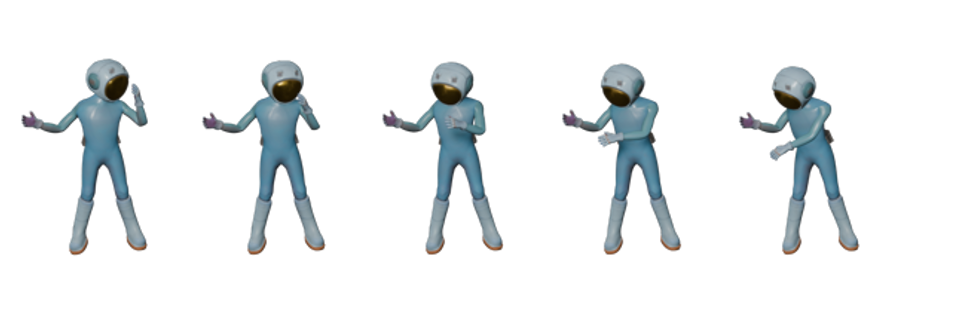}
            \caption{Strutting}
        \end{subfigure}%
        \begin{subfigure}{0.33\linewidth}
        \includegraphics[width=\linewidth, trim = 0 0.5cm 0 0.3cm]{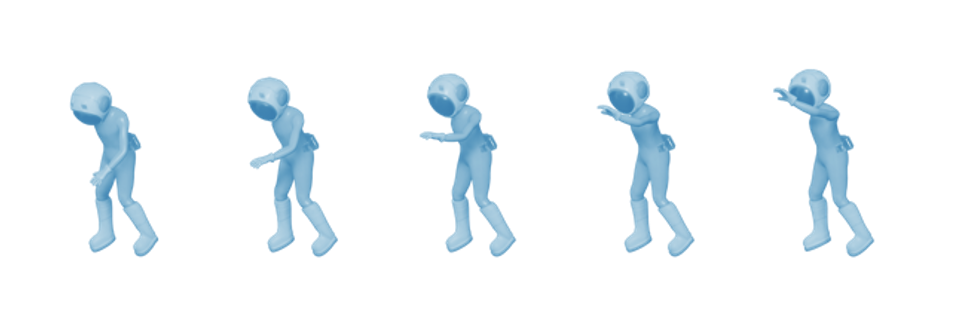}
                    \caption*{}
        \end{subfigure}
        
        \begin{subfigure}{0.33\linewidth}
            \includegraphics[width=\linewidth, trim = 0 0.5cm 0 0.3cm]{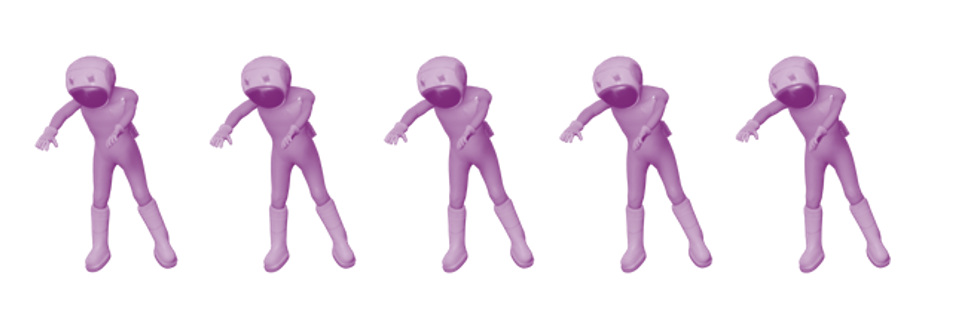}
                    \caption*{}
        \end{subfigure}%
        \begin{subfigure}{0.33\linewidth}
           \includegraphics[width=\linewidth, trim = 0 0.5cm 0 0.3cm]{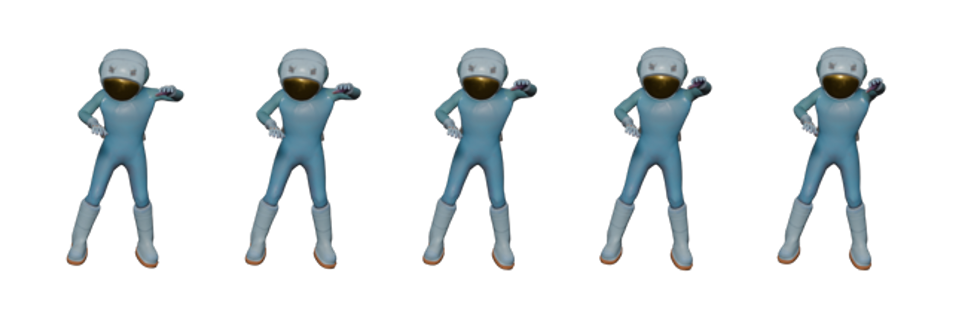}
                    \caption{Zombie}
        \end{subfigure}%
        \begin{subfigure}{0.33\linewidth}
         \includegraphics[width=\linewidth, trim = 0 0.5cm 0 0.3cm]{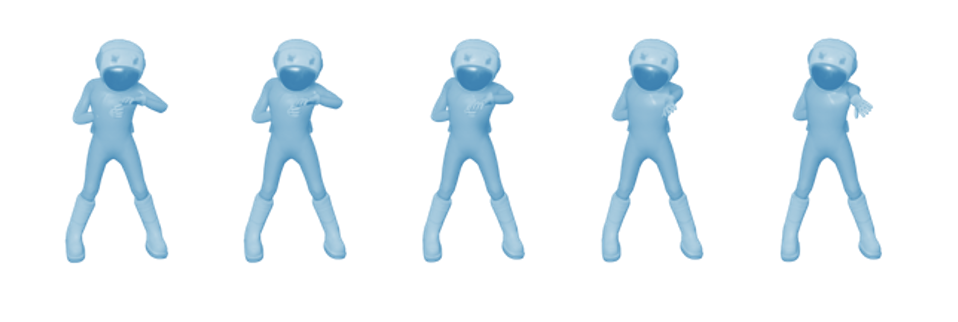}
                    \caption*{}
        \end{subfigure}
        \caption{Style transfer results on \emph{Aberman} dataset. \textbf{Left:} style clips, \textbf{Middle:} transfer results, \textbf{Right:} content clips.}
        \label{fig:style-transfer-aberman}
    \end{figure*}
}

\newcommand{\FigTransitionUnseen}{
    \begin{figure}
        \centering
        \includegraphics[width=\linewidth]{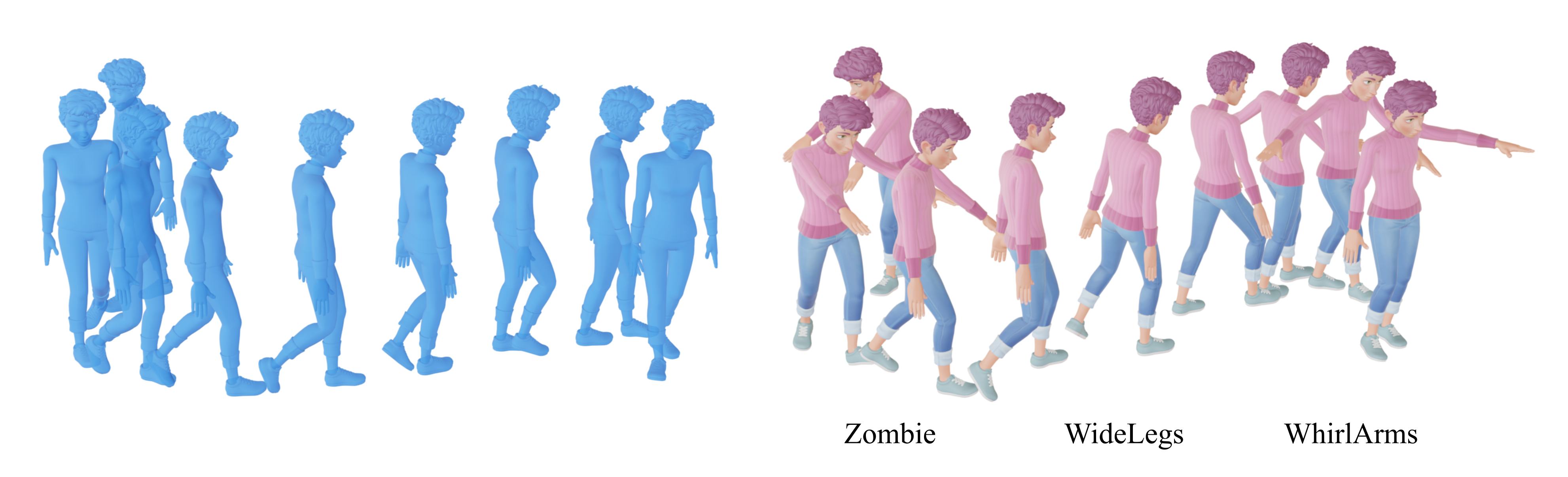}
        \caption{Our method enables switching between multiple styles while following a longer content sequence. The styles used in this motion were unseen during training.}
        \label{fig:transition_unseen}
    \end{figure}
}

\newcommand{\FigStyleNegative}{
    \begin{figure}
        \centering
        \includegraphics[width=\linewidth]{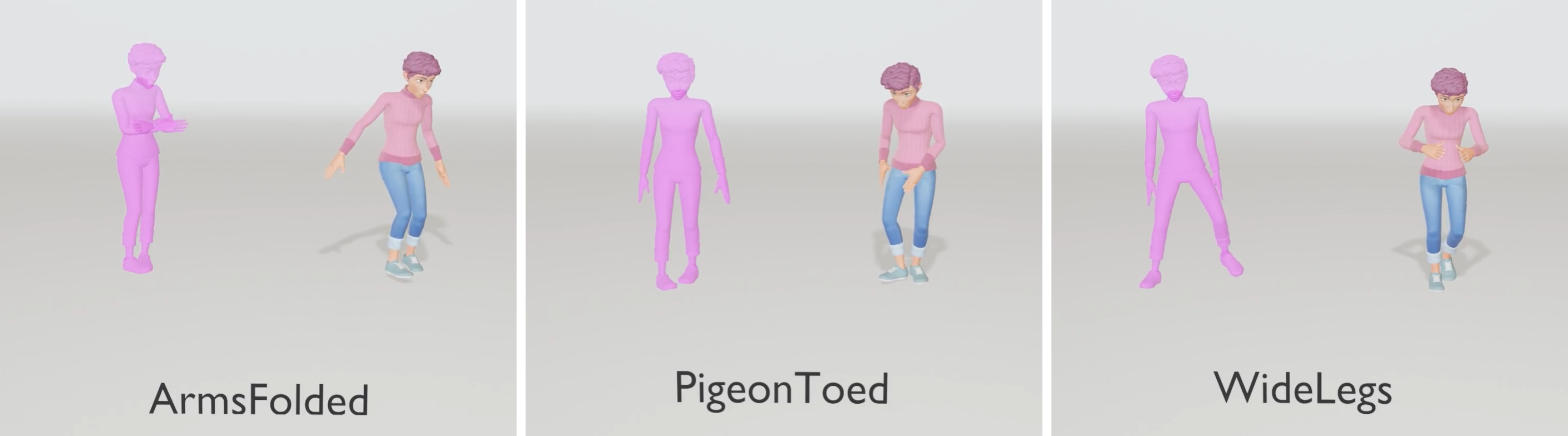}
        \caption{Style inversion. By subtracting the style codes from motion content we obtain a new motion with \emph{inverted} style.}
        \label{fig:style-negative}
    \end{figure}

}

\newcommand{\FigDataAugmentation}{
    \begin{figure}
        \centering
        \includegraphics[width=\linewidth, trim = 0 4cm 0 0]{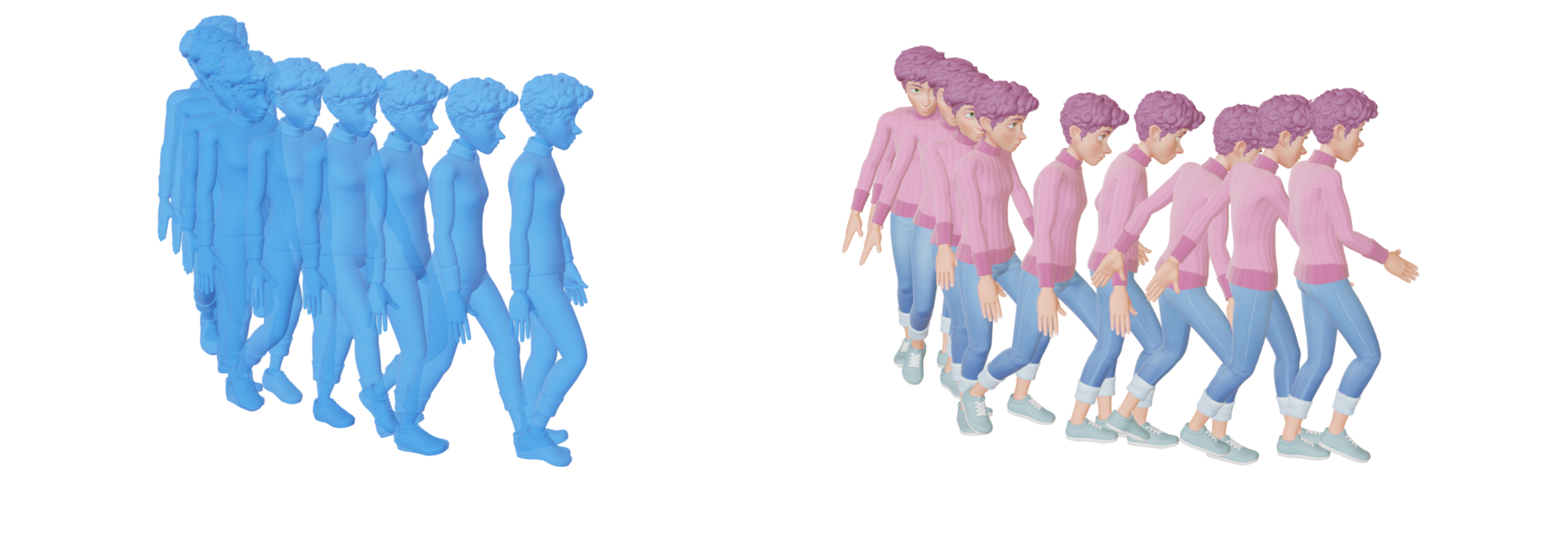}
        \caption{Our pipelines enables data augmentation by following a content motion and sampling random style codes. \textbf{Left:} content motion, \textbf{Right:} result motion.}
        \label{fig:data-augmentation}
    \end{figure}
}

\newcommand{\FigAblationContrast}{
    \begin{figure}
        \centering
        \begin{subfigure}{0.9\linewidth}
            \includegraphics[width=\linewidth, , trim=0 0.4cm 0 0, clip]{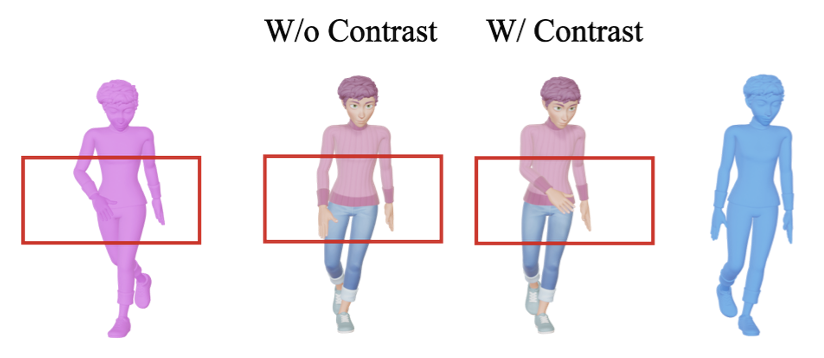}
            \subcaption{Swing Shoulders}
        \end{subfigure}
        \begin{subfigure}{0.9\linewidth}
            \includegraphics[width=\linewidth, trim=0 0.4cm 0 0.9cm, clip]{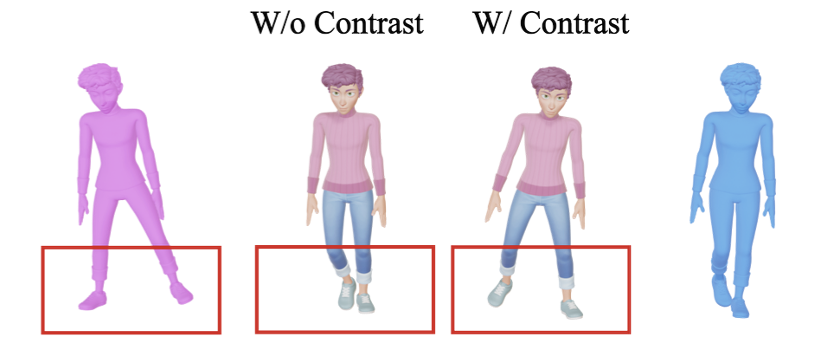}
            \subcaption{Wide Legs}
        \end{subfigure}
        \caption{Contrastive learning improves style-content separation and results in better style transfer. \rev{\textbf{Left (pink):}} style clip, \textbf{Middle (textured):} stylized result of a model trained without (middle left) and with (middle right) contrastive loss, \rev{\textbf{Right (blue):}} content clip.}
        \label{fig:ablation-contrast}
    \end{figure}
}

\newcommand{\FigAblationS}{
    \begin{figure}
        \centering
        \begin{subfigure}{0.85\linewidth}
            \includegraphics[width=\linewidth, trim=0 0.3cm 0 0, clip]{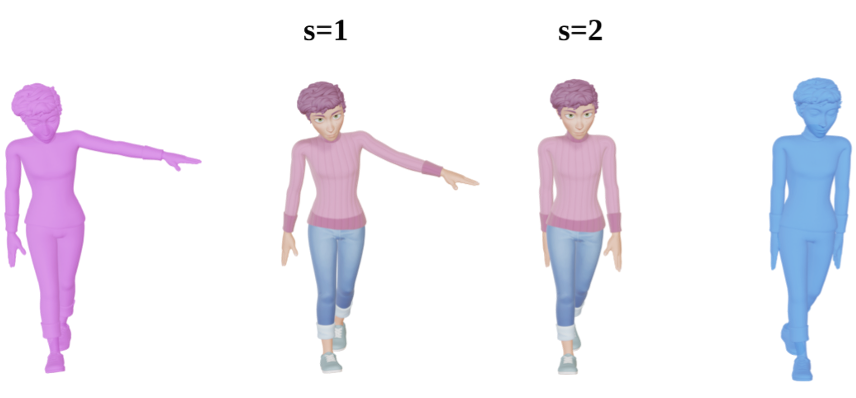}
            \subcaption{Raised Left Arm}
        \end{subfigure}
        \begin{subfigure}{0.85\linewidth}
            \includegraphics[width=\linewidth, trim=0 0.3cm 0 1cm, clip]{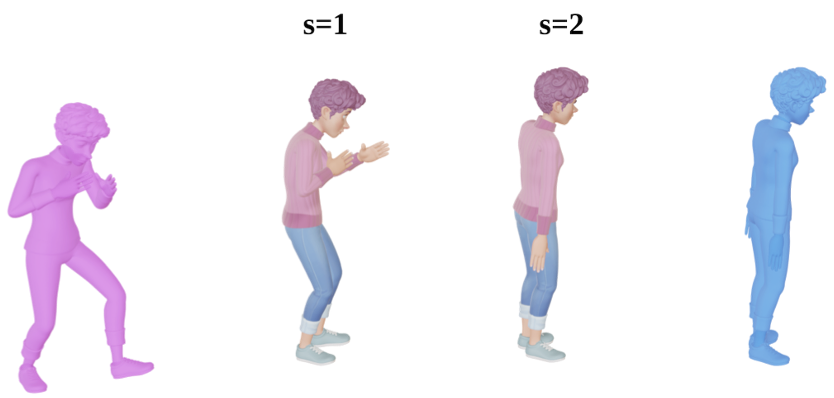}
            \subcaption{Swat}
        \end{subfigure}
        \caption{Ablation on different cut-offs for style-content codebooks in style transfer application. \textbf{Left (pink)}: style clip, \textbf{Middle}: style transfer result, \textbf{Right (blue)}: content clip.}
        \label{fig:ablation-s}
    \end{figure}
}

\newcommand{\TableIanStyles}{
    \begin{table}[tb]
        \centering
        \rev{
        \caption{Baseline comparison on \emph{100STYLE} dataset with LPN-Style\cite{mason2022real}. 
        Numbers in parenthesis show top-5 accuracy.  
        LPN-Style$^{*}$ denotes the experiment excluding 8 broken styles (those with accuracy lower than 10\%).}
        \begin{tabular}{lccc}
        \toprule
         & \multicolumn{3}{c}{Style-acc ($\%$) $\uparrow$  } \\
         \cmidrule(lr){2-4} 
        Method & Test & Unseen style & Fine-tuned\\
        \midrule
        LPN-Style
         & $73.24 (87.50)$ & \text{\ding{55}} & $86.92 (\mathbf{100.0})$\\
        LPN-Style$^{*}$ 
        & $77.53 (91.70)$ & \text{\ding{55}} & $86.92 (\mathbf{100.0})$\\
        Ours 
        & $\mathbf{ 83.20 (95.12) }$ & $\mathbf{ 68.95 (98.83) }$ & $\mathbf{ 96.88} (99.02)$ \\
        \midrule
        & \multicolumn{3}{c}{Content-err (m) $\downarrow$}\\
        \cmidrule(lr){2-4}
        Ours & $0.075 \pm 0.056$ & $0.091\pm0.073$ & $0.079 \pm 0.065$ \\
         \midrule
         & \multicolumn{3}{c}{Cross-classification (\%) $\downarrow$}\\
        \cmidrule(lr){2-4}
        Ours & $0.78$ & $0.00$ & $0.00$\\
        \bottomrule
        \end{tabular}
        \label{tab:100styles}
        }
    \end{table}
}
\newcommand{\TableAberman}{
    \begin{table*}[tb]
        \centering
        \caption{Comparison to \cite{guo2024generative} on \emph{Aberman} and \emph{Xia} datasets. Numbers in parenthesis show top-3 accuracy.}
        \begin{tabular*}{\textwidth}{@{\extracolsep{\fill}} lcccccccc}
        \toprule
         & \multicolumn{3}{c}{Style-acc ($\%$) $\uparrow$  } & \multicolumn{3}{c}{Content-err (m) $\downarrow$} 
         & \multicolumn{2}{c}{Cross-cls ($\%$) $\downarrow$}
         \\
         \cmidrule(lr){2-4} \cmidrule(lr){5-7} \cmidrule{8-9}
        Method & Train & Test & \emph{Xia} styles & Train & Test & \emph{Xia} styles & Train & Test\\
        \midrule
        GenMoStyle & $73.47 (93.36)$ & $76.02 (92.34)$ & 30.77 (76.92) & $\mathbf{0.018 \pm 0.012}$ & $\mathbf{0.019 \pm 0.014}$ & $\mathbf{0.024 \pm 0.018}$ & $7.14$ & $8.16$ \\
        Ours &
        $\mathbf{83.38 (96.88)}$ & $\mathbf{ 80.91 (92.27)} $ & $\mathbf{53.85 (76.92)}$ 
         & $0.028 \pm 0.028$ & $0.029\pm 0.027$ &  $ 0.047 \pm 0.026$ & $\mathbf{3.06}$ & $\mathbf{5.10}$ \\   
        
        \bottomrule
        \end{tabular*}
        \label{tab:Aberman}
    \end{table*}
}

\newcommand{\TableAblation}{
    \begin{table}[ht]
        \centering
        \small
        \caption{Ablation results for contrastive and mutual info losses.}
        \begin{tabular}{lccc}
        \toprule
        Model &  Style-acc (\%) $\uparrow$ & Content-err (m)$\downarrow$ & Rec-err (m) $\downarrow$ \\
        \midrule
        RVQ & 55.86 & $0.076 \pm 0.063$ & $\mathbf{0.029 \pm 0.016}$ \\
        + $\mathcal{L}_{con}$ & 68.95 & $\mathbf{0.074 \pm 0.059}$ & $0.029 \pm 0.017$ \\
        + $\mathcal{L}_{mi}$ & 73.24 & $0.081 \pm 0.065$ & $0.031 \pm 0.020$ \\
        + $\mathcal{L}_{con}$ + $ \mathcal{L}_{mi}$ & \textbf{87.50}  & $0.077 \pm 0.059$ & $0.031 \pm 0.019$ \\
        \bottomrule
        \end{tabular}
        \label{table:ablation}
    \end{table}
}
\section{Introduction}
Virtual characters play a central role in media and entertainment, from animation and video games to immersive experiences.
However, creating realistic and expressive character animation remains a labor-intensive process that often requires extensive manual work from artists, making it both time-consuming and costly.
Recent research in character animation has explored how modern machine learning techniques can reduce repetitive work and automate aspects of the animation pipeline.
One notable problem is \emph{style transfer}, where the \emph{style} of a motion---such as walking happily or angrily---is transferred from one motion clip to another, while preserving the content, or semantic meaning, of the latter motion, which is usually referred to as the \emph{content} clip.

Achieving compelling motion style transfer requires disentangling style from content---a fundamental and challenging problem in motion representation.
While humans naturally exhibit a wide range of walking styles, clearly defining and separating style and content in motion data is non-trivial.
Our goal is to learn a representation that effectively disentangles the content and style from motion clips and allows for transferring styles realistically. 
We approach this problem by interpreting content as the coarse and structural attributes of motion, and style as the finer details that introduce expressive nuances to the movement.
To achieve this separation, we employ Residual Vector Quantized Variational Autoencoders (RVQ-VAEs) \cite{lee2022autoregressive} to learn a coarse-to-fine representation. This is complemented by novel contrastive and mutual information losses that prevent style leakage into content.
We demonstrate that the resulting representation effectively separates style from content and enables style transfer through \emph{Quantized Code Swapping}.
Our approach performs style transfer as an inference-only task, requiring no additional fine-tuning for unseen styles. 
Furthermore, unlike prior works, our model is trained without adversarial or cyclic training, enabling a more stable convergence.

We evaluate our method on multiple motion capture datasets \cite{mason2022real, aberman2020unpaired, xia2015realtime} and show that our representation embeds the content in the initial codebooks and the style in latter ones.
Our experiments show effective style transfer across both seen and unseen stylistic motions, measured via style classifier accuracy, while maintaining the original content, measured via trajectory deviation.
In addition, it enables further capabilities including smooth transitions between styles, content extraction, motion interpolation, and content-style mixing for data augmentation.

In summary, our contributions are as follows:
\begin{itemize}
    \item We learn an interpretable coarse-to-fine representation of motion that disentangles content and style.
    \item We propose a novel strategy by utilizing a mutual info loss to prevent style leakage, and combining contrastive learning with non-differentiable residual codebook learning.
    \item Our framework supports several applications at inference, including style transfer and style transitions over arbitrary-length motions, via swapping and blending operations on the disentangled quantized codebooks.
\end{itemize}
\section{Related Work}

\cite{xia2015realtime} were one of the first to apply style transfer to motion data using a motion database and a KNN search. They first pre-process the database to track the closest neighboring pose for each pose in the dataset. Afterwards, they can perform real-time style transfer relying on a KNN search and simple linear transformation. In the context of deep learning, \cite{aberman2020unpaired} learn separate style and temporal content codes. They use Instance Normalization (IN) and Adaptive Instance Normalization (AdaIN) to remove the original style from the content motion and reintroduce the desired style codes, respectively. Since the model relied heavily on style labels and a discriminator, it struggles to generalize to unseen styles. Both these works also released their annotated styled motion data that have been used by many works since.

\cite{park2021diverse} use a graph convolutional network (GCN) where the pooling and unpooling operations are performed according to the human skeleton. In addition, they learn a mapping network from random noise to a style code to perform style transfer from unseen styles at inference time. Motion Puzzle \cite{jang2022motion} also uses a similar GCN, but trains from one source motion to multiple target motions each to stylize one or more body parts. This lets them combine different styles across body parts during inference. More recently, \cite{tang2024decoupling} used contact information for style transfer via learning separate GCNs to extract contact, trajectory and style information from the source motion. They can then perform style transfer at inference by swapping the style codes passed to the generator transformer. Similarly, the contact timing can also be swapped between the style and content motion.

\textit{100STYLE} dataset\cite{mason2022real} is a locomotion dataset with a combination of 100 different styles and different gaits (walking, running, \& strafing). In addition to introducing the dataset, \cite{mason2022real} train a mixture of expert (MoE) model with a FiLM \cite{dumoulin2018feature, perez2018film} generator and a pre-trained periodic autoencoder \cite{starke2022deepphase} providing learned phase labels. Although their approach works well for seen styles, they require fine-tuning to adapt to unseen styles. \cite{tang2023rsmt} also use an MoE model along with the periodic autoencoder to learn a motion VAE to reconstruct the next frame conditioned on the current frame. The decoder can be conditioned on a style token to perform style transfer. \cite{dai2025towards} further extend the periodic autoencoder to learn body part-wise phase labels. They can then encode these part-wise phase labels into style labels, allowing mixing styles across body parts similar to Motion Puzzle. 
These methods rely on a pretrained phase manifold for motion, a MoE and a motion sampler, constructing a complex training pipeline.

Diffusion models have also been employed for motion style transfer. \cite{raab2024monkey} show that for a pre-trained motion diffusion model \cite{tevet2022human}, the queries and keys in the self-attention blocks correspond to motion \textit{outline} and \textit{motifs} respectively. Using noise inversion and swapping the queries and keys between a content and a style motion clips, they can perform zero-shot style transfer. In contrast, \cite{zhong2024smoodi} learn a ControlNet \cite{zhang2023adding} based style adapter and combine classifier-based and classifier-free \cite{ho2022classifier} guidance to adapt a pre-trained text-to-motion model for stylization. \cite{song2024arbitrary} specifically train a diffusion model for style transfer and first learn a multi-condition extractor that can extract trajectory and remove style from the content motion, and extract the style from the style motion. A latent diffusion model is then trained on these features to stylize the content motion.

While the above-mentioned diffusion models offer promising results, their iterative generative process makes them too slow for real-time applications and for stylizing motion of arbitrary lengths. Hence, \cite{guo2024generative} first learn a motion latent representation using a VAE \cite{kingma2019introduction}. This representation is then disentangled using style and content encoders and a generator. Reconstruction, homo-style alignment, and cycle consistency losses are used to ensure that the new latent space is sufficiently disentangled. Although they can transfer unseen styles, their method requires training multiple models, and losses such as homo-style alignment and cycle consistency can make the training more unstable.

To our knowledge, we are the first work to explore VQ-VAEs \cite{van2017neural} in the context of motion stylization. We show that using a Residual VQ-VAE (RVQ-VAE) \cite{lee2022autoregressive}, we can learn a disentangled latent space for motion style transfer without specialized style and content encoders and perform style transfer for arbitrary long motion sequences in real time.

\section{Method}
\FigFramework

Human motion can be represented as a sequence of $T$ states,
$\mathcal{M} = \{\mathbf{s}_t\}$.
We define the state of the character at timestep $t$ as $\mathbf{s}_t=[\mathbf{p}_t, \mathbf{R}_t, \mathbf{v}_t, \boldsymbol{\omega}_t, \mathbf{h}_t, \mathbf{u}]$, 
where  $\mathbf{p}_t \in \mathbb{R}^{3\times J}, \mathbf{R}_t \in \mathbb{R}^{6 \times J}, \mathbf{v}_t \in \mathbb{R}^{3 \times J}, \boldsymbol{\omega}_t \in \mathbb{R}^{3 \times J}$ represent the position, 6D orientation, linear velocity and angular velocity of each body in local (root) frame, respectively.
$\mathbf{h}_t \in \mathbb{R}^{3}$ is the global height of the root, and $\mathbf{u} \in \mathbb{R}^3$ denotes the global up direction in the local frame.
$J$ is the total number of joints in the skeleton.

Our goal is to learn a motion representation that effectively disentangles the stylistic characteristics of human motion, from its semantic content.
We utilize an RVQ-VAE \cite{yao2024moconvq, lee2022autoregressive} framework that consists of an encoder $\mathcal{E}$, a decoder $\mathcal{D}$ and the latent space is modeled by multiple categorical distributions or codebooks $\mathcal{B}$.
Overview of our framework is demonstrated in Fig. \ref{fig:framework}.
We propose a training strategy in which the initial codebooks capture the content of the motion, while the subsequent codebooks encode the stylistic nuances.
\rev{
Key components of our training strategy include contrastive learning (Sec.~\ref{sec:contrastive}) and mutual information loss (Sec.~\ref{sec:mutual_info}).
This disentangled representation can be utilized at inference time to perform style transfer via Quantized Code Swapping (Sec.~\ref{subsec:latent_code_swapping}).
}

\subsection{Quantized Motion Embedding}
RVQ-VAEs can be trained by sampling a random number of residual codebooks in use, e.g. for each training sample, only the first $n$ number of codebooks $[\mathcal{B}_0, \mathcal{B}_1, ...,\mathcal{B}_{n-1}]$ are employed to reconstruct the data.
This training strategy effectively enables learning a coarse-to-fine representation \cite{yao2024moconvq, lee2022autoregressive}, where the initial codebooks encode coarse information and the subsequent codebooks capture finer details.
Intuitively, the content of a motion clip corresponds to these coarse components of the data, while stylistic variations are expressed through finer nuances.
Under this perspective, RVQ-VAEs provide a natural and effective framework for disentangling style from content in motion.

We train an RVQ-VAE for reconstruction of motion data.
The encoder $\mathcal{E}$ is a 1D convolutional neural network that downsamples the motion sequence and the decoder $\mathcal{D}$ is a 1D deconvolutional neural network that upsamples the motion back to the original sequence length.
A motion sequence $\mathcal{M}$ with $T$ frames is encoded into $K$ latent embeddings $\mathbf{r}_0 = [\mathbf{r}_0^k]_K = \mathcal{E}(\mathcal{M})$.
Then, the first codebook $\mathcal{B}_0$ is used to quantize each embedding in $\mathbf{r}_0$ into the nearest codes $\mathbf{z}_0 = [\mathbf{z}_0^k]_K$,
\begin{equation}
    \mathbf{z}_0^k = \arg \min_{\mathbf{c}_i \in \mathcal{B}_0} ||\mathbf{r}_0^k - \mathbf{c}_i||_2^2 = \mathcal{Q}_{\mathcal{B}_0}(\mathbf{r}_0^k),
\end{equation}
where $\mathcal{Q}_{\mathcal{B}}(.)$ denotes the quantization w.r.t. the codebook $\mathcal{B}$.

The quantized code is used to compute the residual to previous continuous embedding 
\begin{equation}
    \mathbf{r}_i = \mathbf{r}_{i-1}-\mathbf{z}_{i-1} = \mathbf{r}_0-\sum_{j=0}^{i-1} \mathbf{z}_j,
\end{equation}
which is then projected iteratively to the next codebooks,
\begin{equation}
    \mathbf{z}_j^k = \arg \min_{\mathbf{c}_i \in \mathcal{B}_j} ||\mathbf{r}_j^k - \mathbf{c}_i||_2^2 .
\end{equation}

The input to the decoder is prepared by using the first $n$ embedded codes, where $n \leq N$ is sampled randomly during training and $N$ is the maximum number of codebooks.
The final motion $\mathcal{M}'$ is reconstructed by
\begin{equation}
    \mathcal{M}'=\mathcal{D}\left(\sum_{j=0}^{n-1}\mathbf{z}_j\right).
\end{equation}

\subsection{Training}
To train the motion embedding, we use a weighted reconstruction loss on the data,
\begin{equation}
    \mathcal{L}_{rec} = ||w \cdot (\mathcal{M} - \mathcal{M}')||_2^2,
\end{equation}
where $w$ is a weighting vector applied element-wise to the signal to emphasize specific motion features.

To prevent error accumulation along the kinematic chain, we employ a forward kinematics (FK) loss,
\begin{equation}
    \mathcal{L}_{FK}= ||\mathbf{p}_g-\mathbf{p}_g'||_2^2,
\end{equation}
where $\mathbf{p}_g$ are ground truth global positions and $\mathbf{p}_g' = \text{FK}(\mathcal{M}')$, where $\text{FK}(\cdot)$ refers to forward kinematic operation on joint orientations in $\mathcal{M'}$.

To ensure temporal consistency in the generated motion, we use velocity loss, $\mathcal{L}_{vel}$, defined as the mean squared error in the velocities of the generated motion as follows.
\begin{equation}
    \mathcal{L}_{vel}=||\dot{\mathbf{p}}_g - \dot{\mathbf{p}}_g'||_2^2,
\end{equation}
where $\dot{\mathbf{p}}_g$ and $\dot{\mathbf{p}}_g'$ are velocities calculated via finite differences. Lastly, to regularize the motion, we penalize very high accelerations in the output motion via the acceleration loss,
\begin{equation}
    \mathcal{L}_{acc}=||\ddot{\mathbf{p}}_g'||_2^2,
\end{equation}
where $\ddot{\mathbf{p}}_g'$ refers to the finite-difference acceleration of $\mathbf{p}_g'$.

\rev{
To effectively train the codebooks $\mathcal{B}_i$, we use commitment loss and Exponential Moving Average (EMA) update as detailed in Appendix \ref{app:background}.
}
Code reset is also done similar to \cite{lancucki2020robust} for codes that are unused during training to encourage more uniform usage of all codes in the codebook.

\subsection{Contrastive Learning}
\label{sec:contrastive}
\rev{
While the inherent coarse-to-fine structure of RVQ-VAE provides a basic separation of style and content, with initial codebooks encoding coarse content and deeper codebooks capturing the stylistic details, this separation arises purely from reconstruction objectives, and does not achieve a sufficiently robust disentanglement for style transfer applications. 
To enhance this separation, we incorporate a contrastive learning objective that organizes the latent motion embeddings according to style labels, pulling embeddings with the same label closer together and pushing embeddings with different styles farther apart.
We apply the contrastive objective exclusively to the deeper residual codebook embeddings intended to encode the style, leaving the initial content codebooks untouched.
}

\rev{
For this purpose, we adopt the Multi-Pos contrastive loss introduced in \cite{multiposCL} to contrast all pairs of positive and negative samples within a batch.
Let $\mathbf{a}$ denote an anchor embedding sampled from the batch, and $\{\mathbf{b}_1, \mathbf{b}_2, ..., \mathbf{b}_x\}$ denote the remainder.
The Multi-Pos contrastive loss is given by the cross entropy, $H$, between similarity-based distribution of how closely $\mathbf{a}$ matches each $\mathbf{b}$ and target distribution constructed from ground truth style labels.
}
\begin{equation}
    \mathcal{L}_{con} = H\left(\frac{\exp(\mathbf{a}\cdot\mathbf{b}_i/\tau)}{\sum \exp(\mathbf{a}\cdot\mathbf{b}_i/\tau)}, \frac{\mathds{1}_{match(\mathcal{S}(\mathbf{a}), \mathcal{S}(\mathbf{b}_i))}}{\sum \mathds{1}_{match(\mathcal{S}(\mathbf{a}), \mathcal{S}(\mathbf{b}_i))}}\right),
\end{equation}
where $\mathcal{S}(.)$ obtains the style label of a sample and $\tau$ is a tunable hyperparameter. 

\cite{hadjeres2020vector} introduced contrastive learning within a VQ-VAE framework, simply by applying the contrastive objective to the continuous embeddings prior to quantization.
In contrast, our method applies contrastive loss directly to the residual embeddings, i.e. after quantization.
We show that this choice enables backpropagation to update the codes in the latest codebook, without explicitly affecting the gradients of the earlier stages in the pipeline.
This is particularly crucial for our goal, since the style-related contrasts should not affect the initial codebooks which are intended to encode content. 
To show this, note that gradients of the forward path are computed using straight-through method \cite{huh2023straightening}:
\begin{equation}
    \nabla_{\theta} (\mathbf{z}_i) := \nabla_{\theta}(\mathbf{r}_i) 
    \label{eq:straight_through},
\end{equation}
where $\theta$ corresponds to network parameters before codebook $\mathcal{B}_i$.
The gradient of the next residuals will be:
\begin{equation}
    \nabla_{\theta}(\mathbf{r}_{i+1}) = \nabla_{\theta}(\mathbf{r}_i - \mathbf{z}_i) = \nabla_{\theta}(\mathbf{r}_i) - \nabla_{\theta}(\mathbf{z}_i) = 0,
\end{equation}
where the last equality comes from Eq. \ref{eq:straight_through}. The gradient with respect to the codebook is

\begin{equation}
    \begin{aligned}
        \nabla_{\mathcal{B}_i} (\mathbf{r}_{i+1}^k)&=\nabla_{\mathcal{B}_i}(\mathbf{r}_i^k)-\nabla_{\mathcal{B}_i}(\mathbf{z}_i^k)=0 - \nabla_{\mathcal{B}_i}(\mathcal{Q}_{\mathcal{B}_i}(\mathbf{r}_i^k)) \\ 
        &= -\mathds{1}_{match}(\mathbf{z}_i^k, \mathbf{c}_i), \quad \forall \mathbf{c}_i\in \mathcal{B}_i ~.
    \end{aligned}
\end{equation}

\rev{
A second key distinction from \cite{hadjeres2020vector} lies in how the codebooks are optimized. While their method uses the standard VQ objective, we adopt the more robust EMA update (see Appendix \ref{app:background}).
Because EMA updates are applied manually rather than through backpropagation, they break differentiability. Therefore, incorporating the contrastive loss into codebook learning requires careful design.
Although the contrastive loss can update the codebook via backpropagation, EMA is typically performed in the forward step and prior to backpropagation, invalidating the gradients. 
To properly incorporate the contrastive updates, the EMA update must instead be applied after backpropagation, as a second update step.
}

\subsection{Mutual Information Loss}
\label{sec:mutual_info}
The contrastive loss on style labels enables disentanglement of different styles in the residual codebooks.
However, style information can still be stored in the content codebook.
Incorporating a similar content contrastive loss for the content codebooks is infeasible as semantic rich content labels are not always available.
Instead, we propose restricting the model from inferring style labels from the content codebooks.
This can be achieved using a mutual information loss \cite{fano1961transmission, na2019miso}.
Unlike prior work that typically maximize mutual information between given labels and embeddings, we minimize the mutual information between the selected content codes, $z$, and style labels, $l$.
This ensures that no style can be inferred from the content embeddings.
This is formulated as,

\begin{equation}
    \mathcal{L}_{mi} = I({Z}_{content}; \mathcal{S})=\sum_{z\in Z_{content}}\sum_{l \in \mathcal{S}}p(z,l) \log{\frac{p(z,l)}{p(z)p(l)}}.
\end{equation}
To estimate $p(z,l)$ we use Monte-Carlo sampling,
\begin{equation}
    p(z=c_i,l=l_j)\approx \frac{1}{N}\sum_i^N q(z=c_i|r)\cdot \mathds{1}_{match}(l,l_i),
\end{equation}
where $q(z=c_i|r_i)$ denotes the assignment probability of quantization, which can be represented as a one-hot vector. 
However, for better gradient propagation, we replace it with a soft assignment probability,
\begin{equation}
    q(z=c_i|r)=\frac{\exp(-||r-c_i||^2 / \tau)}{\sum_j \exp(-||r-c_j||^2 / \tau)},
\end{equation}
where $\tau$ is a tunable hyperparameter.

\FigFrameworkStyleTransfer

\subsection{Quantized Code Swapping}
\label{subsec:latent_code_swapping}

After learning a quantized motion representation that effectively disentangles style and content into separate codebooks, the style transfer tasks can be easily performed at inference time.

To achieve style transfer, we introduce \emph{Quantized Code Swapping} (see Fig. \ref{fig:framework_style_transfer}).
First, the content clip is encoded using the encoder $\mathcal{E}$ and quantization layers $\mathcal{Q}_{\mathcal{B}_{i<N}}$, yielding the corresponding codes $\mathbf{z}_{i, \text{content}}$.
Similarly, the style clip is encoded to obtain $\mathbf{z}_{i, \text{style}}$.
We then swap the codes after a specified residual layer $s$, replacing those from the content clip with the corresponding codes from the style clip.
The resulting set of codes are then combined and passed through the decoder to produce the final motion,
\begin{equation}
    \bar{\mathcal{M}} = \mathcal{D}\left(\sum_{i=0}^s \mathbf{z}_{i, \text{content}} + \sum_{i=s+1}^{N-1}\mathbf{z}_{i, \text{style}}\right).
\end{equation}
The resulting motion preserves the content of the original clip while adopting the style of the reference style clip.
\section{EXPERIMENTS AND RESULTS}
\FigLatentVis
We evaluate the effectiveness of our method in disentangling style from content in motion clips and demonstrate its applicability across several tasks, including style transfer, style transition, inverse style, motion blending and data augmentation.
Finally, we present a detailed quantitative analysis of style transfer task, along with ablation studies.
For a comprehensive overview of the results, including style transfer and other demonstrations, please refer to the supplementary video.

\subsection{Style-Content Disentanglement}
In this section, we evaluate the disentanglement between style and content in our motion embedding via low-dimensional projection of the residual vectors and by performing applications such as content extraction and style interpolation using interpretable operations.
\label{sec:disentanglement}

\textbf{Low-dimensional projection}.
Fig. \ref{fig:latent_vis} presents a TSNE \cite{van2008visualizing} visualization of the residual embeddings $\mathbf{r}_i$ for each layer, color-coded by the style labels.
As shown in Fig. \ref{fig:latent_vis_a}, even without contrastive learning, the RVQ-VAE architecture already clusters motions of same style close together in the second codebook.
When the contrastive loss is introduced, the separation between styles becomes more distinct (Fig. \ref{fig:latent_vis_b}).
Adding the mutual information loss further strengthens the disentanglement (Fig. \ref{fig:latent_vis_c}).
Notably, the model also exhibits disentanglement for styles that were unseen during training (Fig. \ref{fig:latent_vis_d}).

\FigContentOnly
\textbf{Extracting content}.
Given the residual structure of the embeddings, the content of a motion clip can be readily extracted by decoding using only the content codebooks while discarding the subsequent residual layers. 
Fig. \ref{fig:content_only} shows a comparison between full motion reconstruction and a reconstruction using only the content codes.
As can be seen, the style is effectively removed while the global trajectory, feet placement and semantics remain close to the original motion. Interestingly, since the \emph{100STYLE} dataset is a highly stylized dataset, the neutral motion extracted via our disentanglement keeps its hands extended to the side instead of more closely following the \textit{Neutral} style.

\FigInterpolate

\textbf{Style Interpolation}. Our representation enables interpolating between content motion and stylized motion by scaling the style codes prior to decoding. 
\begin{equation}
    \bar{\mathcal{M}} = \mathcal{D}\left(\sum_{i=0}^s \mathbf{z}_{i} + \alpha \sum_{i=s+1}^{N-1} \mathbf{z}_{i}\right)
\end{equation}
where $0<\alpha<1$ is the interpolation factor. 
As shown in Fig. \ref{fig:interpolate}, intermediate values of $\alpha$ determine the strength of the stylization applied to the content motion. 
In particular, when $\alpha = 0$, the result corresponds to \emph{content extraction}.

\subsection{Style Transfer}
\FigStyleTransferTrain
\FigStyleTransferTest
\FigStyleTransferAberman

Style transfer from a style reference clip onto a content motion clip can be achieved using the \emph{quantized code swapping} technique described in Sec. \ref{subsec:latent_code_swapping}. 
In Fig. \ref{fig:style-transfer-train}, we present results of transferring the styles \emph{Old} and \emph{LeanBack} from \emph{100STYLE} dataset onto two different content clips. 
The generated motion preserves the trajectory of the content motion (shown in blue) while incorporating the stylistic characteristics of the reference clip (shown in pink).
This inference-time style transfer confirms that coarse information in the motion encodes the content while finer details correspond to the style.

Since our learned latent space is both interpretable and disentangled, it enables zero-shot style transfer to new unseen styles. 
As illustrated in Fig. \ref{fig:style-transfer-test}, two novel styles, \emph{WildLegs} and \emph{Zombie}, which were not seen during training are successfully applied to a content clip.

Training our model on \emph{Aberman} dataset which consists of fewer styles (16 styles) on a more general contents further than locomotion, we can also perform style transfer for these motions.
Fig. \ref{fig:style-transfer-aberman} illustrates several examples of style transfer for \emph{Depressed}, \emph{Strutting} and \emph{Zombie} styles.

\subsection{Style Transition} \label{subsec:style_transition}
Given a content motion, we can transition between various styles by concatenating style codes from different style clips along the time dimension before decoding. 
This enables generating long motion sequences and switching between multiple styles.
As seen in Fig. \ref{fig:transition_unseen}, our method can transition between very different styles such as \emph{Zombie}, \emph{WideLegs} and \emph{WhirlArms}, none of which were seen during training. 

The \emph{100STYLE} dataset contains only individual clips for each style, so transitions between different styles are never observed during training. Thanks to the power of the learned representation, we can reliably perform style transitions at inference time.

\FigTransitionUnseen

\subsection{Style Inversion}

In contrast to adding the style codes $z_{i>s}$ during reconstruction, one can subtract them to effectively invert the style of the input motion. 
Fig. \ref{fig:style-negative} demonstrates the results of this operation, revealing that certain styles can be interpreted as the inverse of one another. 
For instance, inverting the \textit{ArmsFolded} style results in the arms being spread apart. 
Similarly, inverting \emph{PigeonToed} and \emph{WideLegs} results in \emph{DuckFeet} and \emph{NarrowLegs} respectively.

Beside being an interesting application, this reflects that our training strategy produces an interpretable latent space that matches our natural intuition of style.

\FigStyleNegative

\subsection{Motion Blending}
Effectively stitching and blending between two motion clips is a key challenge to enable faster animation creation. 
Naive approaches to stitching of two clips often produce discontinuities in the transition.
Using our RVQ-VAE, we can smoothly stitch two motion clips by concatenating the latent codes of the two clips and decoding the result.
The final output seamlessly blends the motions without discontinuities and eliminating the need for manually designed blend functions.

\subsection{Data Augmentation}
As demonstrated in previous works \cite{agrawal2024trajectory, maeda2022motionaug}, data augmentation and bias reduction can significantly enhance motion generation with deep learning models. 
Our interpretable latent space also enables new data augmentation methods.

\textbf{Content interpolation}.
Interpolating content codes, $z_{i\leq s} $ between two different motion clips results in a valid motion with novel content. 
This technique can be used to enhance the diversity of motion trajectories in the dataset. 
As shown in the supplementary video, interpolating between two content codes results in a trajectory that effectively averages the original trajectories.
This approach can reliably introduce new turns and sequences of actions that were not present in the original dataset.

\textbf{Random style selection}
As discussed in Sec. \ref{subsec:style_transition}, our model can generate smooth transitions between styles of different reference clips. 
Even in the absence of a style clip, we can apply multiple styles to a content clip by replacing its style codes, $z_{i>s}$, with different codes randomly selected from our residual codebooks. 
The resulting motion remians smooth while exhibiting multiple styles, enabling natural style transitions that are absent in the original dataset. 
As shown in Fig. \ref{fig:data-augmentation}, the generated motion follows the content from the input clip (on the left), while sequentially exhibiting multiple distinct styles.
These two techniques are orthogonal and can be easily combined to increase both trajectory diversity and style transition variety in the dataset.

\FigDataAugmentation

\subsection{Quantitative analysis}

\TableIanStyles

\TableAberman
In this section, we quantitatively compare our method with the baselines \cite{mason2022real} and \cite{guo2024generative}, in terms of \textit{style preservation accuracy} and \textit{content trajectory deviation}.
We compare against the motion-based supervised setting for \cite{guo2024generative} as it is closest to our inference configuration.
We define \textit{style preservation accuracy}, $A_{S}$ as the classification accuracy of a pretrained style classifier over the generated motion.
\textit{Content trajectory deviation}, $D_C$ measures the deviation of the generated motion's root trajectory from that of the content motion.
For more details on the metric definition and dataset splits, please refer to Appendix \ref{app:metrics}.

\subsubsection{Style Accuracy}
\rev{
In Table \ref{tab:100styles}, we report style accuracy, $A_{S}$, for the test and the unseen style subsets of the \emph{100STYLE} \cite{mason2022real} dataset.
As a baseline, we use the locomotion controller LPN-Style provided by \cite{mason2022real} to generate stylized motion to measure their test $A_{S}$.
In practice, we observed that LPN-Style produced severely broken or incorrectly stylized motion for certain styles, with accuracies falling below 10\%.
To ensure a fair evaluation, we report their performance both including and excluding these problematic styles in the test set.
Our method scores top-1 $A_{S}$ of 83.20\% on the test set, outperforming both versions of LPN-Style.

For unseen styles, our method achieves 68.95\% top-1 and 98.83\% top-5 $A_{S}$, whereas LPN-Style is incapable of zero-shot generalization.
Since LPN-Style cannot generate motion for unseen styles, they must fine-tune their FiLM module \cite{perez2018film} on the new styles.
This fine-tuning improves their performance to 86.92\% $A_S$, while our similarly fine-tuned model can achieve a score of 96.88\%, significantly outperforming LPN-Style.
}

We evaluate our method on the \emph{Aberman} \cite{aberman2020unpaired} and \emph{Xia} \cite{xia2015realtime} datasets which contain more generic stylized motions beyond locomotion.
For training, we used the retargeted motions from \emph{Aberman} provided by GenMoStyle\cite{guo2024generative}.
Our model is trained solely on the \emph{Aberman} training set, but we evaluate it both on the \emph{Aberman} and \emph{Xia} datasets.
Since these datasets contain fewer styles (16 for \emph{Aberman} and 8 for \emph{Xia}), we report top-3 $A_{S}$ rather than top-5 $A_{S}$ used for \emph{100STYLE}.

As reported in Table \ref{tab:Aberman}, our method outperforms GenMoStyle for style accuracy across all subsets and both top-1 and top-3 $A_{S}$.
Notably, while GenMoStyle do not use the \emph{Xia} dataset to train their style and content encoders, their general motion autoencoder is pretrained on all \emph{CMU}\cite{cmu}, \emph{Aberman} and \emph{Xia} datasets.
In contrast, we use the \emph{Xia} dataset exclusively for evaluation.

\subsubsection{Content Error}

We evaluate the content deviation $D_C$ of our method across the three datasets.
Since the \emph{100STYLE} dataset is a locomotion dataset, we observe a higher test deviation of \rev{7.5} cm (see Table \ref{tab:100styles}) compared to 2.9 cm and 4.7 cm on \emph{Aberman} and \emph{Xia} datasets (see Table \ref{tab:Aberman}), respectively.

As \cite{mason2022real} train a locomotion controller rather than a style transfer model, their content deviation error cannot be measured.
On the \emph{100STYLE} dataset, we find that much of the deviation arises from errors in predicted velocities during sharp turns in the content motion.
Since the final motion is calculated by integrating the velocities over time, these velocity errors accumulate and lead to larger deviations toward the end of the sequence.
Operating in a global reference space could potentially improve content preservation; however, we leave this investigation for future work.

As content and style are more closely coupled for non-locomotion data, we see a stronger correlation between better style removal and worse content deviation for such data.
This is corroborated when comparing content deviation and cross-classification rate between our method and GenMoStyle.
The cross-classification rate measures the percentage of times the style classifier classified the generated motion as the style of the content clip.
On \emph{Aberman} and \emph{Xia} datasets, GenMoStyle provides lower content deviation compared to our method.
However, we see that GenMoStyle's generated motion is misclassified as the content style 8\% over test data compared to 5\% for our method.
This suggests that our method is better able to remove the style of the content motion.
As a result, our method removes more of the content clip motion. 
By adjusting the number of content codebooks $s$, one can fine-time the model performance for their preferred balance of style transfer and content preservation.

\subsection{Ablation Studies}

\subsubsection{Effect of contrastive learning}
In this section, we evaluate the effect of contrastive loss, $\mathcal{L}_{cl}$ and mutual information loss, $\mathcal{L}_{mi}$.
As discussed in Sec. \ref{sec:disentanglement}, adding contrastive loss to the training improves the style-content separation. 
This is also confirmed by our experiments on style transfer via quantized code swapping. 
As shown in Fig. \ref{fig:ablation-contrast}, the style transfer result from the model trained with contrastive loss adheres more to the nuances of style clip and shows less leakage of content, compared to the one obtained without contrastive loss. 
For instance, in the \emph{WideLegs} style, the result with contrast opens the legs wide while the one without it keeps the legs close to each other similar to the content clip. 
A similar phenomena can be seen in the \emph{SwingShoulders} style when comparing the arms movement. 

\TableAblation
\FigAblationContrast

Our quantitative measures are consistent with the qualitative results mentioned above.
In Table \ref{table:ablation}, we measure $A_{S}$, $D_C$ and reconstruction error $L2P$ for a base RVQ-VAE, adding each losses individually and training with both losses simultaneously.
Across all four models, we measure similar reconstruction error with only training with RVQ-VAE with the lowest $L2P$.
The difference between our full model and the base RVQ-VAE lies within the standard deviation range, hence, our disentanglement of the learnt space does not result in notably worse reconstruction.

We see that adding each individual loss improves style classification accuracy compared to the base RVQ-VAE.
As we measure $D_C$ after swapping codes, we see that adding \rev{$\mathcal{L}_{con}$} also improves the content error, as it encourages the initial codebooks to contain most of the content information.

In contrast, since $\mathcal{L}_{mi}$ forces the content codes to have no style information, it is moved into later codebooks, resulting in a greater improvement in $A_{S}$ compared to \rev{$\mathcal{L}_{con}$}.
\rev{
However, we notice that adding $\mathcal{L}_{mi}$ results in higher $D_C$ error.
This could be since the MI loss prevents style information leakage to the first codebook, it may also remove trajectory information that is partially style-dependent from the first codebook, resulting in a higher content error.
}
Lastly, the addition of both losses compounds their effect and results in the best style accuracy and a reasonable content trajectory error.

\subsubsection{Choice of swapping cut-off}
In our experiments with basic training of RVQ-VAE without contrastive and mutual info losses, the first codebook captures the content and the second codebook onwards contain information about the motion style. 
This is inline with the intuition that the style is the second greatest feature after content that is represented in the data. 
Therefore we found $s=1$ is a good choice for the cut-off between content and style codebooks.
Fig. \ref{fig:ablation-s} shows style transfer results for $s=1$ and $s=2$, where the latter almost replicates the content motion, indicating that the second codebook already captures most of the stylistic features.
Although one can use only the first and second codebooks for decoding the motion in case of style transfer, we find addition of more residual style codes helps with overall motion quality and reducing artifacts such as foot sliding.

\FigAblationS

\section{Conclusion and future work}
In this work, we frame content-style separation and motion style transfer as a representation-learning problem. 
Exploiting the coarse-to-fine hierarchy of a Residual VQ-VAE, combined with novel disentanglement losses, we obtain a latent space that effectively separates the style from the content.
This representation unlocks a broad range of downstream capabilities such as style transfer, style interpolation, and data-driven augmentation without per-style fine-tuning. 

Qualitatively, our method maintains content motion's trajectory and timing, while adopting the style of the reference clip, even for styles never observed during training. 
Our quantitative analysis further corroborates our qualitative finding that our method successfully transfers styles to a different content motion with a high style accuracy. 
Furthermore, our method is able to perform zero-shot style transfer for unseen styles.

Our method builds on the intuitive interpretation of content and style as coarse features versus finer details.
However, formally defining and isolating these two components remains challenging.
\rev{
Even across existing public datasets, the criteria for what is considered style are inconsistent. For instance, \emph{kicking} is annotated as content in the \emph{Aberman} dataset, but is classified as a style in the \emph{100STYLE} dataset. 
Prior work has adopted various ways to define the style, e.g. via statistical features such as Gram matrices \cite{holden2016deep}, by shift and amplitude of instance normalization in latent space \cite{aberman2020unpaired}, or by separating the non-temporal features through attention masks \cite{tao2022style}. 
In our work, we view the content as coarse global motion structures, and style as the finer-grained local details. For a locomotion-focused dataset such as \emph{100STYLES}, global root trajectory and speed align more naturally with the content, and foot patterns, arm motions and other nuanced features align more closely with style. Similar to prior studies the boundary between content and style remains ambigious, and in practice the annotated style labels majorly influence the style-content distinction.
}

Furthermore, developing better quantitative metrics for style transfer remains an open task.
For example, root trajectory error, a measure of content preservation, can be misleading.
When applying a style like \emph{Drunk}, the ideal output should intentionally alter the original motion's root trajectory, making a low error score a poor indicator of a successful transfer.
One can find similar counter examples for other metrics provided in the literature such as geodesic distance \cite{guo2024generative}.
This is specifically more challenging for non-locomotion movements in which the expected result of style transfer might be ambiguous.

Such ambigious cases also exist in locomotion, particularly in motions with pronounced foot or hip movements. 
As a result, our methods exhibits reduced performance on such examples.
Furthermore, as we integrate local root velocity to obtain the global character trajectory, the generated motion can drift from the content motion over longer motion sequences.
Although our current formulation achieves strong results, it relies on datasets annotated with style labels.
To disentangle unannotated data, one could perform style discovery through unsupervised clustering, and further combine it with our proposed pipeline.
We leave the investigation of these challenges to future research.

Our investigation demonstrates the capabilities of RVQ-VAEs for interpretable motion representation and enabling flexible, inference-time manipulation of the latent space.
Moreover, the simplicity and generality of our framework open up exciting directions for future research, positioning residual quantization as a promising framework for motion reuse, augmentation and transfer.

\section*{Acknowledgment}
The authors would like to thank Dominik Borer for his technical support and Violaine Fayolle and Doriano van Essen for their artistic support during this project.

\bibliographystyle{eg-alpha-doi} 
\bibliography{references}


\clearpage
\appendix
\section{Background}
\label{app:background}
\rev{
VQ-VAEs \cite{van2017neural} are powerful and compact representation models that have demonstrated success in encoding complex data, including human motion \cite{guo2022tm2t}.
However, preserving high quality reconstruction when encoding a large and diverse dataset typically requires the codebook size to grow exponentially, posing challenges for scalability\cite{shannon1959coding, lee2022autoregressive}.
Residual VQ-VAE (RVQ-VAE) addresses this limitation by employing a hierarchical set of codebooks that encode residual errors at multiple levels\cite{lee2022autoregressive}.
Rather than mapping the entire continuous latent vector to a single code drawn from a large codebook, the RVQ mechanism performs sequential refinement using $N$ codebooks $[\mathcal{B}_0, \mathcal{B}_1, \cdots,\mathcal{B}_{N-1}]$. The first codebook quantizes the initial continuous latent vector, after which a residual is computed by subtracting the quantized output from the original latent. The second codebook then quantizes this residual, producing an additional refinement. This hierarchical residual quantization process continues through all $N$ codebooks, and the final latent representation is obtained by summing the quantized outputs from all stages.

Two main losses are involved in training a VQ-VAE. The first is the commitment loss, which encourages the encoder to generate continuous embeddings $\mathbf{r}_i$ that are closer to their corresponding quantized codes $\mathbf{z}_i$. 
\begin{equation}
    \mathcal{L}_{commit}= ||\mathbf{r}_i-\text{sg}(\mathbf{z}_i)||_2^2,
\end{equation}
where $\text{sg}(.)$ is the stop gradient operator.

The second component is a VQ-objective loss that updates the codebook to align with the continuous embeddings assigned to it,
\begin{equation}
    \mathcal{L}_{VQ} = ||\text{sg}(\mathbf{r}_i) - \mathbf{z}_i||^2_2.
\end{equation}
Rather than optimizing this objective directly, we employ an Exponential Moving Average (EMA) update which has been shown to improve the robustness and stability of codebook training~\cite{lancucki2020robust},
\begin{align}
    N_i &\leftarrow \gamma N_i + (1-\gamma)\sum_{batch}\mathds{1}_{match}(\mathcal{Q}_{\mathcal{B}_i}(\mathbf{r}_i), \mathbf{c}_i), \\
    \mu_i &\leftarrow \gamma\mu_i + (1-\gamma) \sum_{batch} \left(\mathbf{r}_i \cdot \mathds{1}_{match}(\mathcal{Q}_{\mathcal{B}_i}(\mathbf{r}_i), \mathbf{c}_i) \right), \\
    \mathbf{c}_i &= \frac{\mu_i}{N_i}, \quad \forall \mathbf{c}_i \in \mathcal{B}_i.
\end{align}
Here, $\mathbf{c}_i$ corresponds to a codeword in codebook $i$ and $\mathds{1}_{match}(\mathbf{x}, \mathbf{y}) = 1$ if $\mathbf{x}=\mathbf{y}$ and 0 otherwise, showing if the specific code was selected for that data. $N_i$ tracks the usage and $\mu_i$ tracks the mean code, while $\gamma$ is a discount factor.
}

\section{Training hyperparameters}
The hyperparameters used for training our RVQ-VAE over different datasets are detailed in Table \ref{tab:hyperparams}. 
\begin{table}[th]
    \centering
        \caption{Training hyperparameters values for reconstruction on \textit{100Styles}\cite{mason2022real} dataset. Values in parenthesis represent configuration used for training on \emph{Aberman}\cite{aberman2020unpaired} dataset.}
    \begin{tabular}{|l|c|}
        \hline
        Hyperparameter & Value \\
        \hline
         Num residual layers & 8 (4)\\
         Num codes per codebook & 512 (256) \\
         Latent size & 256 \\
         Conv feature size & 512 \\
         Learning rate & 1.0e-4 \\
         Max grad norm clip & 1.0 \\
         $\mathcal{L}_{rec}$ coefficient & 1.0 \\
         $\mathcal{L}_{FK}$ coefficient & 0.01 \\
         $\mathcal{L}_{vel}$ coefficient & 0.1 \\
         $\mathcal{L}_{acc}$ coefficient & 0.05 \\
          $\mathcal{L}_{commit}$ coefficient & 0.05 \\
         $\mathcal{L}_{con}$ coefficient & 0.005 (0.05) \\
         $\mathcal{L}_{mi}$ coefficient & 0.02 (0.12) \\
         \hline
    \end{tabular}
    \label{tab:hyperparams}
\end{table}

\section{Metrics}
\label{app:metrics}

We define style accuracy, $A_{S}$, as the accuracy of a style classifier on motion generated after applying style transfer.
We train one classifier with identical architecture as \cite{guo2024generative}.
This consists of four 1D convolutional layers followed by three deconvolutional layers and one linear layer.
The classifiers are trained on all styles in their respective datasets and have test accuracies of 96.87\% over the \emph{Aberman} \cite{aberman2020unpaired} dataset and 98.57\% over the \emph{100STYLE} dataset\cite{mason2022real}.

To quantify how well the content of the original motion is preserved, we measure the deviation of the root trajectory of the generated motion from the root trajectory of the content clip. This is computed as,
\begin{align}
    D_C = \frac{1}{T}\sum_{t = 0}^T |{p^r}_t - \hat{p^r}_t|, 
\end{align}
where ${p^r}_t$ is the position of the content clip root at frame $t$ and $\hat{p^r}_t$ is the generated root position at the same frame.

\section{Dataset details}

Although the \emph{100STYLE} dataset with its 100 unique styles categories is a large enough dataset to train a model for style transfer, Aberman and Xia datasets are much smaller.
In particular, the Aberman dataset has only 193 minutes of motion data with 16 style categories while Xia dataset has 25 minutes of data with 8 style categories.

We report $A_{S}$ and $D_C$ for train, test, and unseen styles subsets of the datasets.
The train subset contains the same clips as the model was trained on.
The test subset contains clips withheld during training but with the same styles seen during training.

For the \emph{100STYLE} dataset, the unseen dataset contains clips with styles that have never been seen by the model before. These styles are selected by choosing the last 10 styles in the alphabetical order.
\rev{
For comparison with LPN-Style \cite{mason2022real}, we retrain our model on the same data split as theirs, where the unseen styles include \textit{HandsInPockets}, \textit{Roadrunner}, \textit{Skip}, \textit{Star}, \textit{WildArms} and the model is trained on the remaining 95 styles.
}

For the Aberman dataset, we split the Aberman into train and test subsets similar to \cite{guo2024generative}. However, we withhold the complete Xia dataset as the unseen subset. Note that some of the styles in the Xia dataset are also present in the Aberman dataset. 
We choose to split this the datasets as such to for consistent comparison with \cite{guo2024generative}.
However, it's worth noting that the original GenMoStyle only uses Xia dataset as the content, never testing on unseen styles.

\section{Per-style class content deviation}

\begin{figure}
    \centering
    \begin{subfigure}{\linewidth}
        \includegraphics[width=\linewidth]{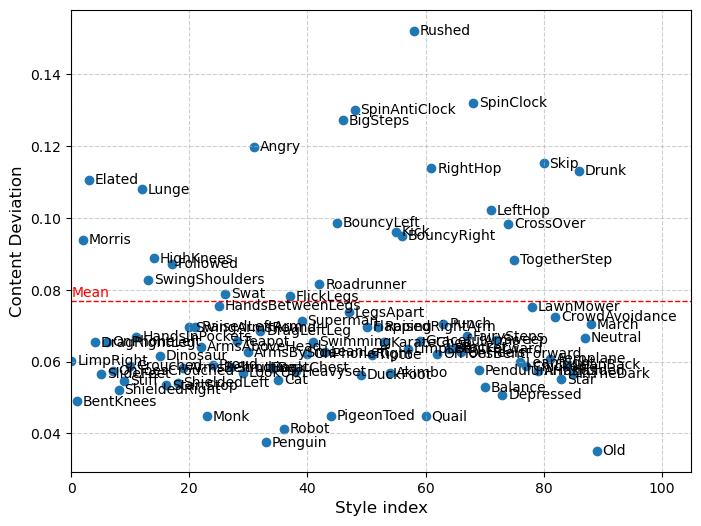}
        \subcaption{Train}
    \end{subfigure}
    
    \begin{subfigure}{\linewidth}
        \includegraphics[width=\linewidth]{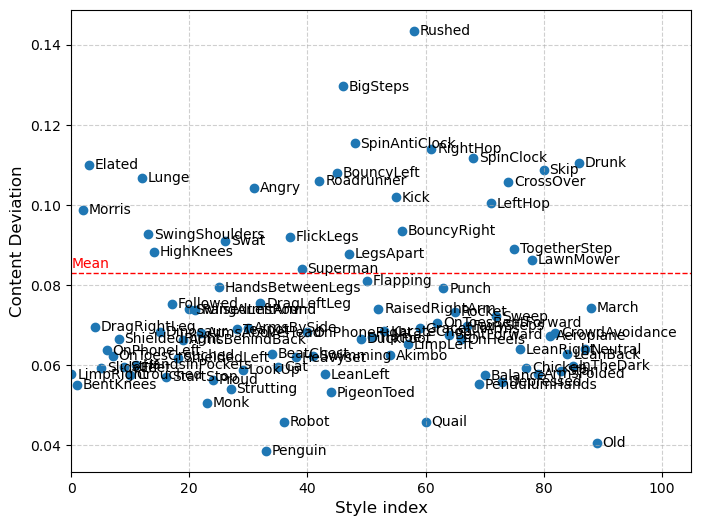}
        \subcaption{Test}
    \end{subfigure}
    \caption{Plotting per class content deviation for train and test subsets of the \textit{100Styles}\cite{mason2022real} dataset.}
    \label{fig:per_style_class_deviation}
\end{figure}

Figure \ref{fig:per_style_class_deviation} plots the content deviation, $D_C$, for the train and test subsets of the \emph{100STYLE}\cite{mason2022real} dataset.
We observe that the distribution of deviations is skewed, with most styles demonstrating a content deviation well below the mean deviation of their respective subsets.
Specifically, of the 90 styles used for training, only 23 styles have an average deviation that exceeds the mean of the training subset.
For the test subset, 25 styles have an average deviation greater than the test subset's mean.
In both training and testing, more energetic styles such as \textit{Rushed}, \textit{Spin}, \textit{Bounce}, and \textit{Hop} have worse content deviation.

\section{Varying number of residual layers}

\begin{table}[t]
\centering
\caption{Rec-err (L2P) (m) vs. different number of codebooks used for reconstruction ($\#s$).}
\begin{tabular}{|c|c|c|c|c|}
\hline
$\#s$ &  RVQ-8-512 & RVQ-8-256 &RVQ-5-512 & RVQ-3-512 \\
\hline
1 & 0.100 & 0.101 & 0.107 & 0.097 \\
2 & 0.061 & 0.065 & 0.068  & 0.074 \\
3 &  0.052 & 0.056 & 0.056 & \textbf{0.067} \\
4 & 0.046 &  0.051 & 0.051 & -- \\
5 & 0.042 & 0.048 & \textbf{0.048} & -- \\
6 & 0.039 & 0.045  & --  & -- \\
7 & 0.037 & 0.043 & -- & -- \\
8 & \textbf{0.036} & \textbf{0.041} & -- & --\\
\hline

\end{tabular}
\label{table:rec-numcodebooks}
\end{table}
Using the residual VQ-VAE architecture, one can use different number of codebooks for reconstructing a motion. Increasing the number of codebooks gradually improves the reconstruction error, as shown in Table \ref{table:rec-numcodebooks}. However, the improvement gets less and less significant as the number of codebooks increase. This aligns with the intuition of coarse-to-fine representation where the later codebooks only add finer details to the constructed motion. 
Models in Table \ref{table:rec-numcodebooks} follow the naming convention of \emph{RVQ-N-X} where $N$ is the total number of codebooks and $X$ refers to the size of each codebook.
Ablation on number of codes in the codebook shows small degrade in the performance when reducing the codebook size.

\end{document}